\documentclass[journal,twoside,web]{ieeecolor}
\usepackage{tmi}
\usepackage{cite}
\usepackage{amsmath,amssymb,amsfonts}
\usepackage{algorithmic}
\usepackage[ruled,linesnumbered]{algorithm2e}
\usepackage{graphicx}
\usepackage{textcomp}
\usepackage{booktabs}
\usepackage{amsmath}
\usepackage[colorlinks,linkcolor=blue]{hyperref}
\usepackage[normalem]{ulem} 
\usepackage{colortbl}
\usepackage{bm}
\definecolor{frenchblue}{rgb}{0.0, 0.45, 0.73}

\usepackage{overpic}
\usepackage{overpic} 
\usepackage{color,}
\usepackage{mathtools} 
\usepackage{currfile}
\usepackage{multirow}

\definecolor{turquoise}{cmyk}{0.65,0,0.1,0.3}
\definecolor{purple}{rgb}{0.65,0,0.65}
\definecolor{dark_green}{rgb}{0, 0.5, 0}
\definecolor{orange}{rgb}{0.8, 0.6, 0.2}
\definecolor{red}{rgb}{0.8, 0.2, 0.2}
\definecolor{darkred}{rgb}{0.6, 0.1, 0.05}
\definecolor{blueish}{rgb}{0.0, 0.3, .6}
\definecolor{light_gray}{rgb}{0.7, 0.7, .7}
\definecolor{pink}{rgb}{1, 0, 1}
\definecolor{greyblue}{rgb}{0.25, 0.25, 1}

\newcommand{\loss}[1]{\mathcal{L}_\text{#1}}
\newcommand{\expect}{\mathbb{E}}

\newcommand{\numfrequencies}{L}
\newcommand{\transpose}{{\operatorname{T}}}

\renewcommand{\paragraph}[1]{\vspace{.5em}\noindent\textbf{#1}.}

\usepackage{lipsum}

\usepackage{blindtext}

\newcommand{\image}{I}
\newcommand{\C}{\hat{\boldsymbol{C}}} 
\newcommand{\gt}{\text{gt}} 

\newcommand{\params}{\boldsymbol{\theta}}
\newcommand{\near}{{t_n}}
\newcommand{\far}{{t_f}}
\newcommand{\radiance}{\mathbf{c}}

\newcommand{\ray}{\mathbf{r}}

\newcommand{\origin}{\mathbf{o}}
\newcommand{\dir}{\mathbf{d}}

\newcommand{\density}{\sigma}

\newcommand{\kostas}[1]{{\color{Bittersweet} {[\bf Kosta: #1]}}}
\newcommand{\andrew}[1]{{\color{BlueViolet} {[Andrew: #1]}}}
\newcommand{\vitto}[1]{{\color{OrangeRed} {[Vitto: #1]}}}
\newcommand{\JB}[1]{{\color{OliveGreen} {[Jon: #1]}}}
\newcommand{\tom}[1]{{\color{RoyalPurple} {[Tom: #1]}}}
\newcommand{\pratul}[1]{{\color{Emerald} {[Pratul: #1]}}}
\newcommand{\at}[1]{{\color{blueish}#1}}
\newcommand{\AT}[1]{{\color{blueish}{\bf [Andrea: #1]}}}
\newcommand{\At}[1]{\marginpar{\tiny{\textcolor{blueish}{#1}}}}
\newcommand{\al}[1]{\textbf{\color{orange}[AL: #1]}}
\renewcommand{\kostas}[1]{}
\renewcommand{\andrew}[1]{}
\renewcommand{\vitto}[1]{}
\renewcommand{\JB}[1]{}
\renewcommand{\tom}[1]{}
\renewcommand{\pratul}[1]{}
\renewcommand{\at}[1]{}
\renewcommand{\AT}[1]{}
\renewcommand{\At}[1]{}
\renewcommand{\al}[1]{}

\newcommand{\norm}[1]{\left\lVert#1\right\rVert}

\def\BibTeX{{\rm B\kern-.05em{\sc i\kern-.025em b}\kern-.08em
    T\kern-.1667em\lower.7ex\hbox{E}\kern-.125emX}}
\begin{document}
\title{Efficient Deformable Tissue Reconstruction \\ via Orthogonal Neural Plane}
\author{Chen Yang, Kailing Wang, Yuehao Wang, Qi Dou, Xiaokang Yang, Wei Shen
\thanks{This work was supported in part by the National Key R\&D Program of China 2022YFF1202600, in part by the National Natural Science Foundation of China under Grant 62176159, in part by the Natural Science Foundation of Shanghai 21ZR1432200, and in part by the Shanghai Municipal Science and Technology Major Project 2021SHZDZX0102. \textit{(Corresponding author: Wei Shen.)}}
\thanks{C. Yang, K. Wang, X. Yang and W. Shen are with the MoE Key Laboratory of Artificial Intelligence, AI Institute, Shanghai Jiao Tong University, Shanghai 200240, China (e-mail:\{ycyangchen, wangkailing151, xkyang, wei.shen\}@sjtu.edu.cn).}
\thanks{Y. Wang and Q. Dou are with Dept. of Computer Science and Engineering, The Chinese University of Hong Kong, Hong Kong 999077 (e-mail:yhwang21@cse.cuhk.edu.hk;~qidou@cuhk.edu.hk)}
}

\maketitle

\begin{abstract}

Intraoperative imaging techniques for reconstructing deformable tissues in vivo are pivotal for advanced surgical systems. 
Existing methods either compromise on rendering quality or are excessively computationally intensive, often demanding dozens of hours to perform, which significantly hinders their practical application.
In this paper, we introduce Fast Orthogonal Plane (Forplane), a novel, efficient framework based on neural radiance fields (NeRF) for the reconstruction of deformable tissues. 
We conceptualize surgical procedures as 4D volumes, and break them down into static and dynamic fields comprised of orthogonal neural planes. This factorization discretizes the four-dimensional space, leading to a decreased memory usage and faster optimization. 
A spatiotemporal importance sampling scheme is introduced to improve performance in regions with tool occlusion as well as large motions and accelerate training.
An efficient ray marching method is applied to skip sampling among empty regions, significantly improving inference speed.
Forplane accommodates both binocular and monocular endoscopy videos, demonstrating its extensive applicability and flexibility.
Our experiments, carried out on two in vivo datasets, the EndoNeRF and Hamlyn datasets, demonstrate the effectiveness of our framework. In all cases, Forplane substantially accelerates both the optimization process (by over 100 times) and the inference process (by over 15 times) while maintaining or even improving the quality across a variety of non-rigid deformations. This significant performance improvement promises to be a valuable asset for future intraoperative surgical applications. The code of our project is now available at \href{https://github.com/Loping151/ForPlane}{https://github.com/Loping151/ForPlane}.

\end{abstract}
\section{Introduction}
\label{sec:introduction}
\IEEEPARstart{R}{econstructing} deformable tissues from endoscopy videos in surgery is a crucial and promising field in medical image computing. High-quality tissue reconstruction can assist surgeons in avoiding critical structures, \textit{e.g.}, blood vessels and nerves, as well as improve the observation of lesions such as tumors and enlarged lymph nodes~\cite{https://doi.org/10.1111/iju.14038, BIANCHI2020e669}. Moreover, tissue reconstruction significantly contributes to creating a virtual surgical training environment~\cite{LANGE200061}. Such detailed reconstructions not only facilitate skill acquisition and expedite the learning curve for endoscopists but also generate substantial training data. This data is crucial for developing virtual reality (VR) and augmented reality (AR) training modules in surgical education, as well as for enhancing the learning algorithms of surgical robots~\cite{9807505, long2022robotic}.
However, achieving precise identification and visualization of these structures and lesions remains a challenge for existing methods. Moreover, the extensive training and inference time required by these methods hinders their practical application during surgery.
This paper aims to address these challenges and seize the opportunities by focusing on high-quality real-time reconstruction of deformable tissues in both monocular and binocular endoscopy videos, providing valuable insights for future intraoperative applications.



Previous methods~\cite{maier2014comparative, mahmoud2018live, semmler20163d, luegmair2015three} primarily focus on static surgical scenes and disregard the presence of diverse surgical instruments in endoscopy videos. These methods rely on simultaneous localization and mapping (SLAM) techniques, generating individual meshes for each frame to reconstruct surgical procedures. However, they fail to account for non-rigid transformations in deformable tissues and potential occlusions caused by surgical instruments. As a result, their reconstruction performance lacks fidelity, contains gaps, and does not realistically represent the surgical environment. These limitations significantly hinder their applicability in data simulation and intraoperative assistance.

\begin{figure*}[ht]
\centerline{\includegraphics[width=0.95\textwidth]{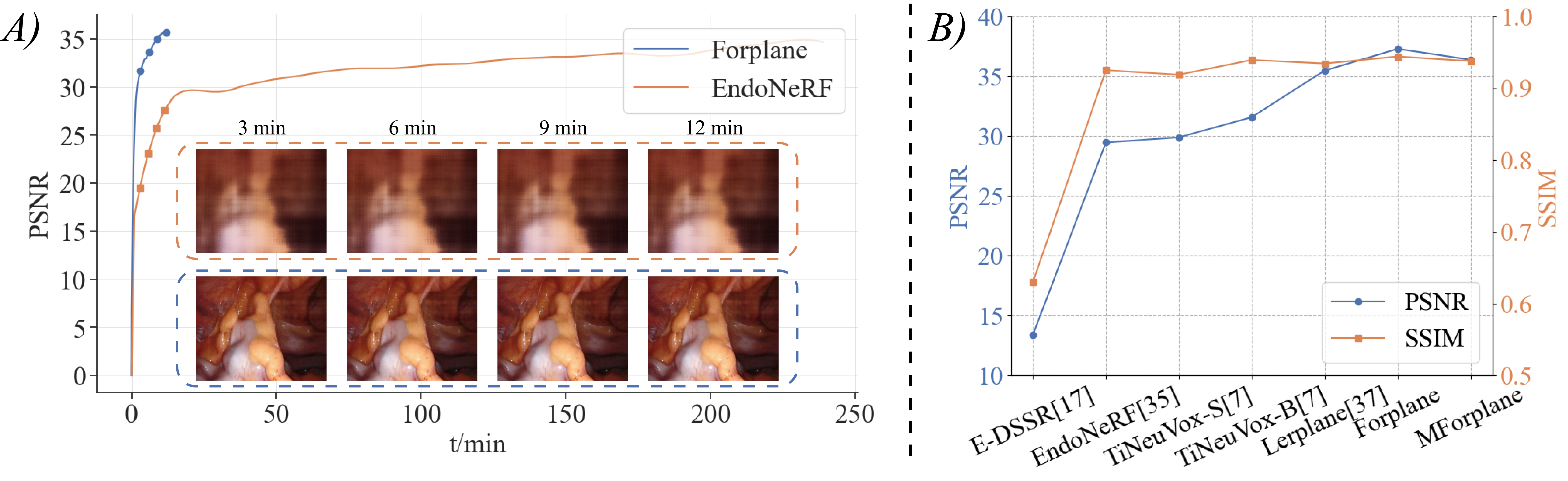}}
\caption{A) The speed-quality comparison between our Forplane and EndoNeRF~\cite{endonerf}. 
We present the performance of the two methods in terms of PSNR and provide insights into their training times. Furthermore, we showcase the reconstruction results obtained from both methods at various training time intervals, specifically at 3, 6, 9, and 12 minutes. These results clearly demonstrate the significant convergence speed exhibited by Forplane.
B) The metrics (PSNR, SSIM) of 7 different methods on the $\textit{pushing tissues}$ scene from EndoNeRF dataset~\cite{endonerf}. 
Both metrics exhibit a positive correlation, indicating that higher values shows better performance.
}
\label{fig: figcurve}
\end{figure*}

Recently, the integration of implicit scene representations and differentiable volume rendering has exhibited remarkable efficacy in capturing complex scenarios.
A notable example is neural radiance fields (NeRF)~\cite{nerf}, a seminal work in neural rendering that introduced neural implicit fields for continuous scene representations. NeRF has demonstrated exceptional performance in tasks such as high-quality view synthesis and 3D reconstruction across diverse contexts~\cite{xie2022neural}. Furthermore, NeRF and its variants have showcased diverse applications in medical imaging, encompassing but not restricted to medical imaging segmentation~\cite{khan2022implicit}, deformable tissue reconstruction~\cite{endonerf, lerplane}, radiograph measurement reconstruction~\cite{li20213d, corona2022mednerf}, and organ shape completion~\cite{yang2022implicitatlas, ruckert2022neat}.

A notable advancement in reconstructing deformable tissue is EndoNeRF~\cite{endonerf}, a recent approach that has demonstrated exceptional capabilities in 3D reconstruction and deformation tracking of surgical scenes within the context of robotic surgery. EndoNeRF~\cite{endonerf} employs a canonical neural radiance field in conjunction with a time-dependent neural displacement field to effectively model deformable tissues using binocular captures within a single viewpoint configuration. However, despite its impressive achievements in reconstructing deformable tissues, EndoNeRF~\cite{endonerf} faces computational challenges due to intensive optimization processes. The optimization process for EndoNeRF~\cite{endonerf} typically requires tens of hours to complete, as each pixel generated necessitates a substantial number of neural network calls. This computational bottleneck significantly restricts the widespread adoption of such methods in surgical procedures.
Additionally, its performance heavily relies on precise binocular depth estimation, further impeding its broader applications.

To overcome the challenges mentioned above, we propose a novel approach called Forplane (Fast Orthogonal Plane), which delivers efficient and high-quality deformable tissue reconstruction and offers rapid training and inference with endoscopy videos. Forplane integrates space-time decomposition with neural radiance fields to achieve rapid and accurate reconstruction of deformable tissues during surgical procedures. By treating surgical procedures as 4D volumes, with time as an orthogonal axis to spatial coordinates, Forplane discretizes the continuous space into static and dynamic fields. Static fields are represented by spatial planes, while dynamic fields are represented by space-time planes.
We utilize limited resolution features to represent these planes, allowing for bilinear interpolation when querying features at any spatio-temporal point. This design markedly reduces computational costs compared to MLP-reliant methods~\cite{park2021nerfies, pumarola2021d, endonerf}, cutting down complexity from O$(N^4)$ to O$(N^2)$.
Additionally, the static field facilitates information sharing across neighboring timesteps, addressing the limitations imposed by restricted viewpoints. 
Furthermore, drawing inspiration from the observation that certain tissues exhibit more frequent deformations, we develope an importance sampling method, which strategically higher sampling probability towards tissues that are either occluded by surgical tools or exhibit more extensive motion range.

A preliminary version of Forplane was introduced in Lerplane~\cite{lerplane}, where its ability to achieve rapid reconstruction of deformable tissues on binocular endoscopy videos was demonstrated. 
\textbf{1) Advancements in Rendering and Speed}: In this extended version, we have made notable advancements on rendering quality, inference speed and adaptability.
Specifically, we develop an efficient ray marching algorithm to guide the rendering procedure, significantly improving inference speed and rendering quality.
\textbf{2) MForplane for Monocular Videos}: Considering that many surgical procedures employ monocular devices, we present MForplane, a version of Forplane adapted for monocular videos.
MForplane performs effectively on monocular sequences alone with minimal performance drop, though this is a more challenging task.
\textbf{3) Comprehensive Evaluation with Hamlyn Datasets}: In addition to validating our results with the EndoNeRF dataset~\cite{endonerf}, we have expanded our evaluation to include the public Hamlyn dataset~\cite{hamlyn1,hamlyn2}. This dataset offers more complex sequences that encompass various challenging factors such as intracorporeal scenes with weak textures, deformations, motion blur, reflections, surgical tools, and occlusions. The inclusion of this dataset allows us to assess the performance of Forplane in more diverse and realistic surgical scenarios.
Fig.~\ref{fig: figcurve} shows that Forplane excels in its significantly faster optimization, demonstrating superior quantitative and qualitative performance in 3D reconstruction and deformation tracking within surgical scenes in comparison to preceding methods. 
Fig~\ref{fig: our performance a} demonstrates the superior reconstruction quality of Forplane.
These results herald significant potential for future intraoperative applications.

The remainder of this paper is organized as follows. Section II provides a concise review of the literature pertinent to our domain. Section III delineates the principal components of our Forplane framework. Section IV details the experimental results, validating the efficacy of our approach. In Section V, we elucidate the mechanisms underlying Forplane's capacity for efficient reconstruction of deformable tissues and outlines future research directions. Finally, Section VI synthesizes our findings and contributions into a conclusion.

\section{Related Work}
\subsection{Surgical Scene Reconstruction}
Numerous endeavors have been undertaken to reconstruct deformable tissues within surgical scenes. 

\subsubsection{Non-implicit Representations}
There have been numerous noteworthy non-implicit approaches in the realm of surgical scene reconstruction~\cite{schmidt2023tracking}. Earlier SLAM-based studies such as~\cite{song2017dynamic, zhou2019real, zhou2021emdq} leveraged depth estimation from stereo videos and fuse depth maps in 3D space for reconstruction. However, these methods either neglected the presence of surgical tools or oversimplified the scenes by presuming them to be static. Later advancements, such as SuPer~\cite{li2020super} and EDSSR~\cite{edssr}, proposed frameworks for stereo depth estimation with tool masking, and SurfelWarp~\cite{gao2019surfelwarp} performed single-view 3D deformable reconstruction. These approaches relied heavily on deformation tracking via a sparse warp field, which compromised their efficacy when encountering deformations that exceed simple non-topological changes.

\subsubsection{Novel Implicit Representations}
Implicit representations like NeRF~\cite{nerf} have made remarkable contributions to medical imaging.
In contrast to non-implicit representations, which encode discrete features or signal values directly, implicit representations use generator functions that associate input coordinates with their respective values within the input space~\cite{molaei2023implicit}.
Recently, EndoNeRF~\cite{endonerf} emerged as a promising solution. It utilized NeRF with tool-guided ray casting, stereo depth-cueing ray marching, and stereo depth-supervised optimization, yielding high-quality non-rigidity reconstruction. However, EndoNeRF optimized an entire spatial temporal field, resulting in significant time and resource consumption. 
Lerplane~\cite{lerplane} factorizes the scene into explicit 2D planes of static and dynamic fields to accelerate optimization, but the inference speed is insufficient to meet clinical demands. 
Therefore, for effective inter-operative application, more intensive efforts must be dedicated towards speeding up the process without sacrificing the reconstruction performance.

\begin{figure}[t]
\centerline{\includegraphics[width=0.9\columnwidth]{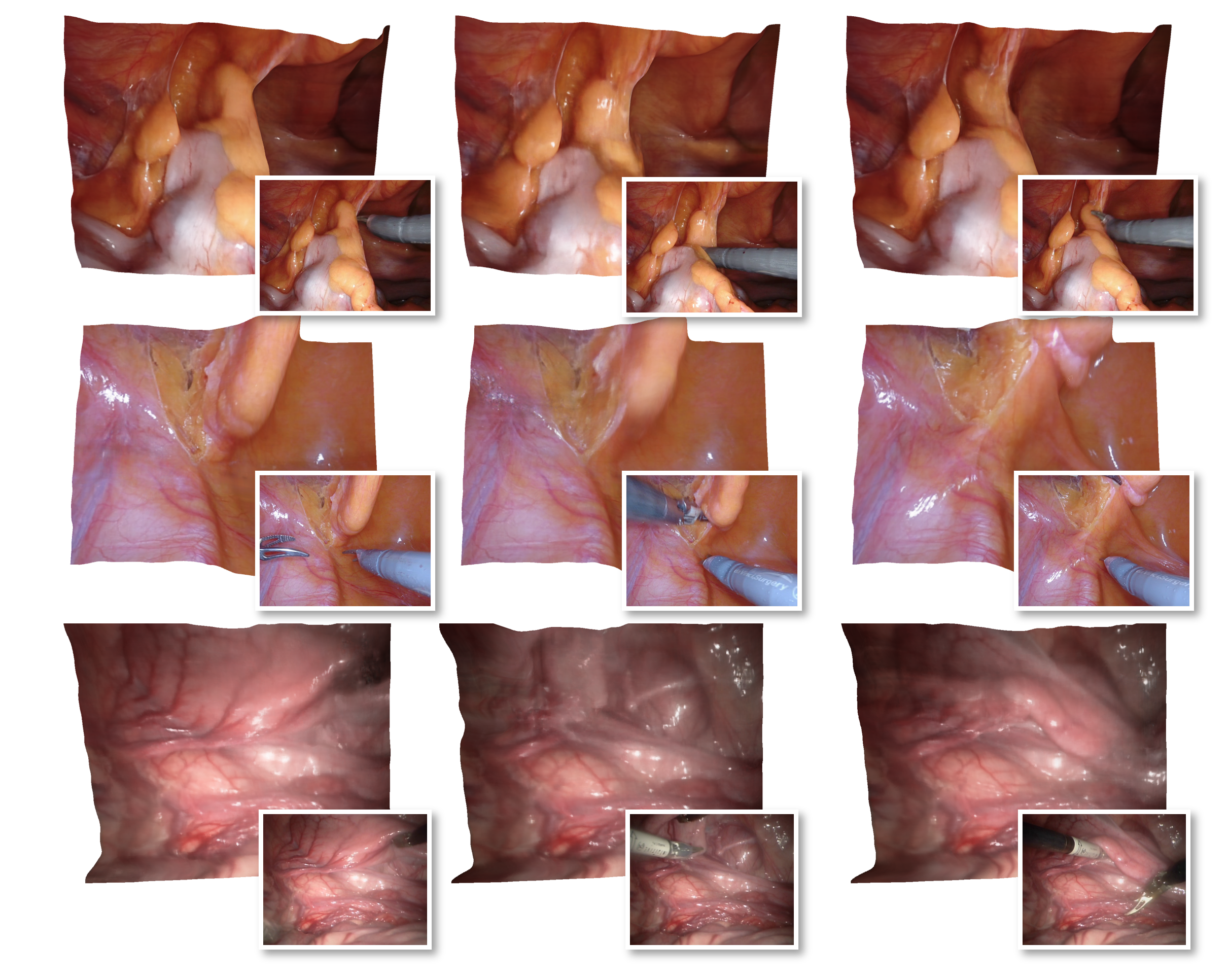}}
\caption{Reconstruction results of deformable tissues. We show the deformable tissues reconstructed by Forplane among surgical procedures with the corresponding captured images in the lower-right corner.}
\label{fig: our performance a}
\end{figure}

\subsection{Implicit Representations in Medical Imaging}
Implicit representations have yield significant contributions in other realms of medical imaging. Some methods leverage NeRF to perform visualization of medical images. For instance, Li et al.~\cite{li20213d} used NeRF to acquire 3D ultrasound reconstructed spine image volume to assess spine deformity. MedNeRF~\cite{corona2022mednerf} rendered CT projections given a few or even a single-view X-ray utilizing NeRF and GAN. NeAT~\cite{ruckert2022neat} used a hybrid explicit-implicit neural representation for tomographic image reconstruction, showing better quality than traditional methods. 
CoIL~\cite{sun2021coil} leveraged coordinate-based neural representations for estimating high-fidelity measurement fields in the context of sparse-view CT.
EndoNeRF~\cite{endonerf}, applied dynamic nerf~\cite{pumarola2021d} to perform deformable surgical scene reconstruction, showing promising performance on stereo 3D reconstruction of deformable tissues in robotic surgery. 
Some methods use implicit functions to do shape reconstruction. ImplicitAtlas~\cite{yang2022implicitatlas} proposed a data-efficient shape model based on templates for organ shape reconstruction and interpolation. 
Fang et al.~\cite{fang2022curvature} introduced a curvature-enhanced implicit function network for high-quality tooth model generation from CBCT images. 
Raju et al.~\cite{raju2022deep} presented deep implicit statistical shape models (DISSMs) for the 3D delineation of medical images.
For segmentation, IOSNET~\cite{khan2022implicit} used neural fields to create continuous segmentation maps which converge fast and are memory-efficient. 
TiAVox~\cite{zhou2023tiavox} used a time-aware attenuation voxel approach for sparse-view 4D DSA reconstruction. NeSVoR~\cite{10015091} modeled the underlying volume as a continuous function to perform slice-to-volume reconstruction. Reed et al. devised a reconstruction pipeline that utilizes implicit neural representations in conjunction with a novel parametric motion field warping technique to perform limited-view 4D-CT reconstruction of rapidly deforming scenes~\cite{reed2021dynamic}. Schmidt et al. introduced RING~\cite{schmidt2022recurrent} which estimates flow efficiently and KINFlow~\cite{schmidt2022fast} which enables a prior-free estimation of deformation, both using implicit neural representation and graph-based model.
\section{Materials and Methods}

We focus on the task of efficiently reconstructing deformable tissues from both monocular and binocular endoscopy videos. Mathematically, we represent a surgical procedure as a 4D volume denoted by $\mathbf{V}_{4d}$, with dimensions $H\times W\times D \times T$. Here, $H, W, D$ represents the 3D space of the scene, and the $T$ represents the time dim, assumed to be orthogonal to the 3D space. Unlike previous methods that treat the surgical procedure as independent static 3D volumes per time step $\{\mathbf{V}_{3d}^{1}, \mathbf{V}_{3d}^{2},...,\mathbf{V}_{3d}^{T}\}$ or use a time-dependent displacement field $\mathbf{G}_{\boldsymbol\phi}$ to model the tissue deformations, we factorizes the 4D volumes into a static field $\mathbf{V}_{3d}^{s}$ and a dynamic field $\mathbf{V}_{3d}^{d}$ composed of 2D neural planes. This factorization significantly enhances convergence speed and improves the representational capacity.

\begin{figure*}[ht]
\centerline{\includegraphics[width=\textwidth]{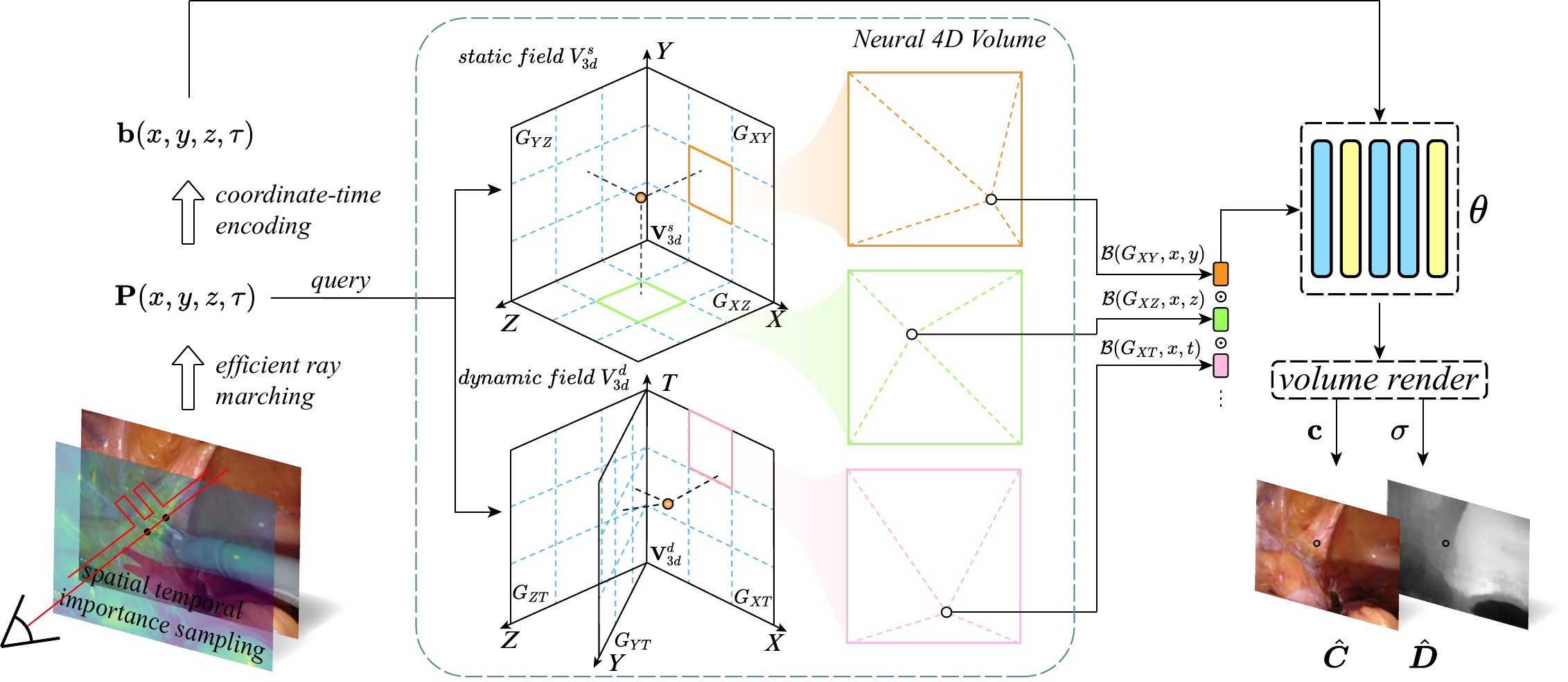}}
\caption{The rendering pipeline of Forplane. Forplane begins with spatiotemporal importance sampling (Sec.~\ref{method: Spatiotemporal Importance Sampling}), which assigns higher sampling probabilities to tissue pixels occluded by tools or with extensive motion. Following the selection of the pixel (ray) to be rendered, an efficient ray marching algorithm (Sec.~\ref{method: efficient ray marching}) samples spatial points along the ray near the tissue surface. The coordinates and time embeddings of sampled points are used to query their corresponding features from multi-scale orthogonal feature planes (\textit{e.g.}, $G_{XY}$) via linear interpolation (Sec.~\ref{method: Neural Forplane Representation for Deformable Tissues}). 
For simplicity, single-resolution planes exemplify these multi-scale neural planes. The features from different planes and scales are fused using element-wise multiplication in Forplane. To enhance temporal information, the fused plane features are concatenated with positional encoded coordinate and time embeddings (Sec.~\ref{method: Coordinate-Time Encoding} and Fig.~\ref{fig: MLP structure}), and then passed through a lightweight MLP~$\params$. Finally, a volume rendering scheme is applied to generate predicted color and depth values for each selected ray.}
\vspace{-1.5em}
\label{fig: pipline}
\end{figure*}

In this section, we begin with reviewing the key techniques used in NeRF~\cite{nerf} and EndoNeRF~\cite{endonerf} (Sec. \ref{method: NeRF and EndoNeRF}). Subsequently, we introduce our novel and efficient orthogonal plane representation for surgical scenes (Sec. \ref{method: Neural Forplane Representation for Deformable Tissues}).
To reconstruct deformable tissues at any time step, we first employ a novel spatiotemporal sampling algorithm to identify high-priority tissue pixels and generate corresponding rays (Sec. \ref{method: Spatiotemporal Importance Sampling}). Next, we design one efficient ray marching method to generate discrete samples along the selected rays (Sec. \ref{method: efficient ray marching}). These samples are then used to retrieve corresponding features from orthogonal planes with linear interpolation.
The obtained features, along with the spatiotemporal information of the samples, are fed into a lightweight MLP that predicts radiance and density for each sample (Sec. \ref{method: Coordinate-Time Encoding}). Finally, we apply standard volume rendering to render the accumulated color and depth along the selected rays. 
To improve the spatiotemporal smoothness and decomposition, we design various regularization strategies on $\{\mathbf{V}_{3d}^{s}, \mathbf{V}_{3d}^{d}\}$ as well as the lightweight MLP (Sec. \ref{method: Optimization}).
We present an optimization strategy for Forplane that diminishes its reliance on binocular depth so that it can work with monocular scopes (Sec. \ref{method: Forplane with Monocular Inputs}).
An illustration of the overall framework is presented in Fig.~\ref{fig: pipline}.

\subsection{NeRF and EndoNeRF}
\label{method: NeRF and EndoNeRF}
We provide a brief overview of the pipeline of NeRF~\cite{nerf} and EndoNeRF~\cite{endonerf}.
Given a set of posed images $\{\mathbf{I}_i\}_{i=1}^N$, NeRF represents a 3D scene using volume density and directional emitted radiance for each point in space with a coordinate-based neural network $\params$. To render one image $\mathbf I_{i}$, a ray casting process is performed for each pixel, where a ray $\mathbf{r}(t) = \mathbf{x}_{o} + t\mathbf{d}$ is projected through the camera pose. Here, $\mathbf{x}_o$ is the camera origin, $\mathbf{d}$ is the ray direction, and $t$ denotes the distance of a point along the ray from the origin. 
NeRF uses a positional encoding $\gamma(\cdot)$ to map the coordinates $\mathbf{x}$ and viewing direction $\dir$ into a higher dimensional space:
\begin{equation}
    \gamma(\mathbf{x})\!=\!\Big[ \sin(\mathbf{x}), \cos(\mathbf{x}), \ldots, \sin\!\big(2^{\numfrequencies-1} \mathbf{x}\big), \cos\!\big(2^{\numfrequencies-1} \mathbf{x}\big) \Big]^\transpose \,. \label{eq:posenc}
\end{equation}
\normalsize
The process of casting a ray through the scene and generating discrete samples along the ray is termed ray marching.
The color $\C(\ray)$ for a specific ray $\ray(t)$ is computed using volume rendering, which involves integrating the weighted volumetric radiance within the near and far bounds $\near$ and $\far$ of the ray:
\begin{align}
\C(\ray) = \int_\near^\far \!\!\!
w(t)
\cdot
\underbrace{\radiance(\ray(t), \dir)}_\text{radiance}~dt
\label{eq:volume_render}
\end{align}
The integration weights $w(t)$ for volume rendering are given by:
\begin{align}
w(t) = \underbrace{\exp \left( -\int_{t_n}^{t} \density(\ray(s)) \, ds \right )}_\text{visibility of $\ray(t)$ from $\origin$} \:\: \cdot \!\!\!\! \underbrace{\density(\ray(t))\vphantom{\int_{t_n}^{t}}}_\text{density at $\ray(t)$}
\label{eq:weights}
\end{align}
The training of $\params$ is supervised by an L2 photometric reconstruction loss.

EndoNeRF~\cite{endonerf} employs a canonical radiance field $F_\Theta(\mathbf{x}, \dir)$ and a time-dependent displacement field $G_\Phi(\mathbf{x}, \tau)$ to represent the surgical scene. The time-dependent displacement field maps the input space-time coordinates $(\mathbf{x}, \tau)$ to the displacement between point $\mathbf{x}$ at time step $\tau$ and its corresponding point in the canonical field.
For any given time step $\tau$, the radiance and density at $\mathbf{x}$ can be obtained by querying $F_\Theta(\mathbf{x} + G_\Phi(\mathbf{x}, \tau), \mathbf{d})$. 
EndoNeRF employs the same positional encoding algorithm to map the input coordinates and time step into Fourier features before feeding them into the networks.

\subsection{Fast Orthogonal Plane Representation}
\label{method: Neural Forplane Representation for Deformable Tissues}
The surgical procedure consists of a series of consecutive frames, each representing a separate scene. Notwithstanding the mutable nature of tissue transformations over time, a significant portion of the tissue structure demonstrates continuity across consecutive frames. Observing this, we propose a method that efficiently capitalizes on the time-invariant components of the tissue structures to construct a static field $\mathbf{V}_{3d}^{s}$. 
The static field is designed to encapsulate invariant tissue structures across these frames, thereby enabling the reuse of static components without necessitating data duplication for each temporal step. This design significantly reduces the memory requirement by obviating the need to repeatedly store information pertaining to the static aspects of the scene in every frame. 
In addressing the time-aware deformations, we have formulated a dynamic field, denoted as $\mathbf{V}_{3d}^{d}$. This field is uniquely tasked with capturing the deviations from the static field. This focused approach allows the dynamic field to circumvent the need for a comprehensive reconstruction of the entire scene, thereby substantially diminishing the computational burden.
To build the static field, 
we adopt the orthogonal space planes (\textit{i.e.}, $G_{\textit{XY}}$, $G_{\textit{YZ}}$ and $G_{\textit{XZ}}$) to represent the static components among the surgical procedure. The use of orthogonal space planes has been proven effective and compact in various static scene reconstruction methods~\cite{tensorf, instantngp}.
As for the $\mathbf{V}_{3d}^{d}$, it is designed to capture the time-dependent appearance. Since the time dimension is orthogonal to the 3D space, it is natural to employ space-time planes (\textit{i.e.}, $G_{\textit{TX}}$, $G_{\textit{TY}}$ and $G_{\textit{TZ}}$) to represent the dynamic field. 
Each space plane has dimensions of $N \times N \times D$, and each space-time plane has dimensions of $N \times M \times D$, where $N$ and $M$ denote the spatial and temporal resolutions, respectively, and $D$ represents the size of the feature stored within the plane.

To render one tissue pixel $p_{ij}$ in a specific time step $\tau$, we first cast a ray $\mathbf{r}(t)$ from $\mathbf{x}_o$ to the pixel. We then sample spatial-temporal points along the ray, obtaining their 4D coordinates. We acquire a feature vector for a point $\mathbf{P}(x, y, z, \tau)$ by projecting it onto each plane and using bi-linear interpolation $\mathcal{B}$ to query features from the six feature planes:
\begin{align}
    \mathbf{v}(x,y,z,\tau) = & \mathcal{B}(G_{\textit{XY}}, x, y) \odot \mathcal{B}(G_{\textit{YZ}}, y, z) \notag \\  & \cdots \mathcal{B}(G_{\textit{YT}}, y, \tau) \odot \mathcal{B}(G_{\textit{ZT}}, z, \tau), \label{eq:interpolate}
\end{align}
where $\odot$ represents element-wise multiplication, inspired by~\cite{fridovich2022plenoxels, cao2023hexplane, fridovich2023k}. The fused feature vector $\mathbf{v}$ is then passed to a tiny MLP $\params$, which predicts the color $\radiance(\ray(t), \dir)$ and density $\sigma(\ray(t))$ of the point. Finally, we leverage the Eq.~\ref{eq:volume_render} to get the predicted color $\C$. Inspired by the hybrid representation of static fields~\cite{instantngp}, we build $\{\mathbf{V}_{3d}^{s}, \mathbf{V}_{3d}^{d}\}$ with multi-resolution planes. 

The factorization of surgical procedure brings three main benefits: 
1) Significant acceleration: Existing methods for reconstructing surgical procedures using pure implicit representations require traversing all possible positions in space-time, resulting in high computational and time complexity. In contrast, querying a Forplane is fast and efficient, involving only several bilinear interpolations among feature planes and a vector matrix product. This reduces the computational cost from O$(N^4)$ to O$(N^2)$. Moreover, this factorization lightens the burden on the MLP, enabling the use of lighter MLP architectures, 
comprising just 2 fully connected layers, each with 64 channels. In contrast, EndoNeRF~\cite{endonerf} employs a complex MLP structure, processing the positional encoding of the input location $\gamma(x)$ through 8 fully connected ReLU layers, each with 256 channels.
2) Sharing learned scene priors: The static field in Forplane facilitates information sharing across different time steps, improving reconstruction performance by incorporating learned scene priors.
3) Modeling arbitrary deformation: Compared with displacement field used by~\cite{du2021neural, li2021neural, park2021nerfies, tineuvox, endonerf} which struggles with changes in scene topology, the dynamic field can easily model complex tissue deformations, allowing for more accurate reconstruction in surgical scenes.

\subsection{Spatiotemporal Importance Sampling}
\label{method: Spatiotemporal Importance Sampling}
Tool occlusion during robotic surgery poses significant challenges in accurately reconstructing occluded tissues due to their infrequent representation in the training set. As a result, different pixels encounter varying levels of difficulty in the learning process, exacerbating the complexity of tissue reconstruction. Furthermore, the presence of stationary tissues over time leads to repeated training on these pixels, yielding minimal impact on convergence and diminishing overall efficiency \cite{yang2023nerfvs}. To tackle these challenges, we devise a novel spatiotemporal importance sampling strategy. 
It prioritizes tissue pixels that have been occluded by tools or exhibit extensive motion ranges by assigning them higher sampling probabilities.

In particular, we utilize binary masks $\{\boldsymbol{M}_i\}_{i=1}^T$ and temporal differences among frames to generate sampling weight maps $\{\boldsymbol{W}_i \}_{i=1}^T$. These weight maps represent the sampling probabilities for each pixel/ray, drawing inspiration from EndoNeRF~\cite{endonerf}.
One sampling weight map $\boldsymbol{W}_i$ can be determined by:
\begin{align}
    \boldsymbol{W}_i &= \min(\max_{\substack{i-n<j\\<i+n}} (\|\ \boldsymbol{I}_i\odot\boldsymbol{M}_i - \boldsymbol{I}_j\odot\boldsymbol{M}_j \|_1)/3, \alpha)\odot\bm{\mathit\Omega}_i, \notag \\ 
    \bm{\mathit\Omega}_i &= \beta(T\boldsymbol{M}_i/ {\sum\limits_{i=1}^T\boldsymbol{M}_i}),
\end{align}
where $\alpha$ is a lower-bound to avoid zero weight among unchanged pixels, $\bm{\mathit\Omega}_i$ specifies higher importance scaling for those tissue areas with higher occlusion frequencies, and $\beta$ is a hyper-parameter for balancing augmentation among frequently occluded areas and time-variant areas. 
By unitizing spatiotemporal importance sampling, Forplane concentrates on tissue areas and speeds up training, improving the rendering quality of occluded areas and prioritizing tissue areas with higher occlusion frequencies and temporal variability.

\subsection{Efficient Ray Marching}
\label{method: efficient ray marching}
The process of volume rendering greatly benefits from the precise sampling of spatiotemporal points, particularly around tissue regions.
Both EndoNeRF \cite{endonerf} and the original version Lerplane \cite{lerplane} introduce specialized sampling schemes to enhance point accuracy. EndoNeRF's approach involves a stereo depth-cueing ray marching module that utilizes stereo depth data to guide point sampling near tissue surfaces. However, this method is limited in scenarios lacking accurate stereo depth information. Conversely, our original version Lerplane \cite{lerplane} employs a sample-net for surface-focused sampling. While effective, this approach requires extensive sampling across the pipeline, leading to reduced inference speeds.

To address this issue, we design an indicator grid that directs the Forplane's focus towards the tissue surface. This method is inspired by the grid representation method~\cite{fridovich2022plenoxels, nerfacc, instantngp, tensorf} used in static scene reconstruction, and it serves to optimize ray marching. The indicator grid functions as a binary map which is cached and updated throughout the training process, signifying empty areas. This low cost grid allows for the early termination of the marching process, based on the transmittance along the ray.



For optimization efficiency, we integrate the indicator grid update with Forplane's standard training. We start with the assumption that all space is dense, and the indicator grid is fully occupied. As we introduce a series of spatial temporal points to Forplane, we pass a small batch sampled from them to the tiny MLP $\params$ to obtain the corresponding density, updating the indicator grid accordingly.
Initially, the optimization process is somewhat slow. However, after several iterations, the indicator grid becomes capable of identifying the density distribution within the 4D volume. This capability significantly reduces the need for unnecessary point sampling, which in turn greatly enhances the speed of training.


Our low cost indicator grid allows for efficient ray marching during both the optimization and inference stage. Leveraging the pre-learned scene context distribution allows us to ignore points with minimal contribution to rendering results and sample points near the surface more accurately. This change improves rendering quality, especially for deformable tissues. Compared to the original version~\cite{lerplane}, our novel efficient ray marching enables faster training and substantially improves rendering speed, approximately 5 times faster (\textbf{\textit{1.73~fps}} vs. \textbf{\textit{0.38~fps}}), representing a significant advancement for intraoperative use.

\subsection{Coordinate-Time Encoding}
\label{method: Coordinate-Time Encoding}
In Section \ref{method: NeRF and EndoNeRF}, it is illustrated that previous methods utilize a positional encoding function $\gamma(\cdot)$ to map coordinates and viewing directions into a higher-dimensional space. The positional encoding of the viewing direction is exploited to capture the view-dependent appearance, which operates on the assumption of a stationary environmental light source. However, this modeling method proves ineffective within the context of an in vivo sequence. The primary reason for this inefficacy is the dynamic nature of the light source, which moves in conjunction with the camera, bound by specific constraints. Consequently, the modeling of the view-dependent appearance becomes arbitrary, which leads to inaccurate scene reconstruction.

\begin{figure}[t]
\centerline{\includegraphics[width=0.8\columnwidth]{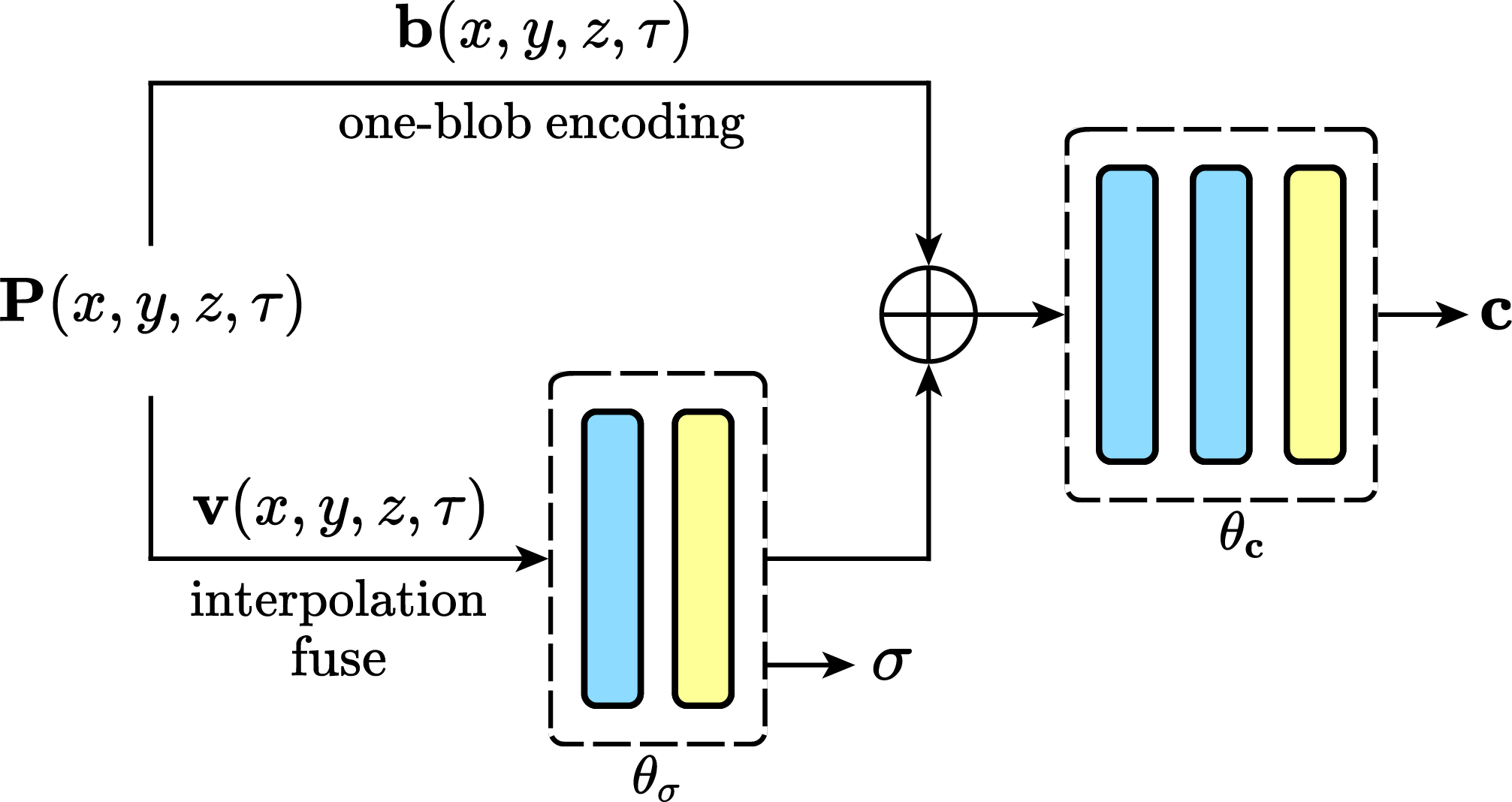}}
\caption{The procedure of Coordinate-Time Encoding. We pass positional encoded coordinate and time embeddings to enhance the temporal information. The tiny MLP ($\params$) consists of the sigma net $\theta_\sigma$ and the color net $\theta_{\mathbf c}$. The blue and yellow boxes denote MLP layers with and without ReLU activation.}
\label{fig: MLP structure}
\end{figure}

Rather than inputting positionally-encoded viewing information, we propose an enhancement on spatiotemporal information, as shown in Fig.~\ref{fig: MLP structure}. This enhancement involves concatenating interpolated features in Eq.~\ref{eq:interpolate} with positionally encoded coordinates and temporal embeddings. This new approach ensures a more precise representation of the scene and a more accurate reflection of the dynamic conditions inherent in an in vivo sequence.
In this work, we use one-blob~\cite{oneblob} encoding separately on each of the four coordinate values. 
The formulation of one-blob encoding is:

\begin{equation}
    \mathbf{b}(s) = g_{s, \xi^2}(\frac{i-0.5}{k}), i = 1, 2, 3\dots k,
\end{equation}
where $g_{s, \xi^2}$ is a Gaussian kernel with mean value $s$ and variance $\xi^2$. 
The encoding function maps $s \in \mathbb{R}$ into a higher dimensional space $\mathbb{R}^k$. 
The encoding along with the fused features $\mathbf{v}$ from feature planes is input to the MLP $\params$, which predicts $\sigma$ and $\mathbf c$ of each point. Then we utilize Eq.~\ref{eq:volume_render} to render the expected color $\hat{\boldsymbol{C}}$ and depth $\hat{\boldsymbol{D}}$ of one specific ray. 


{
{
\begin{algorithm}[t]
\small
\caption{The Rendering Procedure\label{algo:viewcoveragealgo}}
\SetKwInOut{Input}{Input}
\SetKwInOut{Output}{Output}
\SetKwComment{Comment}{/* }{ */}
\SetKwInOut{Define}{Functions}
\Input{A specific ray $\mathbf r(t) = (\mathbf x_o +t\mathbf d )$, 4 levels of the static field and the dynamic field $\{\mathbf{V}_{3d}^{s}$, $\mathbf{V}_{3d}^{d}\}^4$, the indicator grid $\mathbf G_{o}$ and the tiny MLP $\boldsymbol\theta$}
\Output{The expected color $\hat{\boldsymbol{C}}$ and depth $\hat{\boldsymbol{D}}$ of $\mathbf r(t)$}
$\boldsymbol{\mathcal P}_u \leftarrow $Uniformly sampled points \\
$\boldsymbol{\mathcal P}\leftarrow \mathbf G_o(\boldsymbol{\mathcal P}_u)$ to get sample points around the surface\\
\For{$\mathbf P(x, y, z, \tau)$ in $\boldsymbol{\mathcal P}$}
{   
    $\mathbf v(x, y, z, \tau) \leftarrow \mathbf 1$ to initialize\\
    \For{$G, ij$ in $\operatorname{zip}(\{\mathbf V^s_{3d}, \mathbf V^d_{3d}\}^4, \{(xy, xz, x\tau, yz, y\tau, z\tau\})$}
    {
        $\mathbf v(x, y, z, \tau) \leftarrow \mathbf v(x, y, z, \tau) \odot \mathcal B(G, i, j)$,\\
        ~$\mathcal B$ \text{is the bi-linear interpolation on $G$}
    }
    $\text{features} \leftarrow \operatorname{concat}(\mathbf b(x, y, z, \tau), \mathbf v(x, y, z, \tau))$ \\
    $(\mathbf c_i, \mathbf \sigma_i) \leftarrow \boldsymbol\theta (\text{features})$
}
$\hat{\boldsymbol C} \leftarrow \int_{t_n}^{t_f} w(t) \cdot \radiance(\mathbf r(t), \mathbf d)~dt$, $\hat{\boldsymbol D} \leftarrow \int_{t_n}^{t_f} w(t)\cdot t~dt$, $w(t) = \exp \left( -\int_{t_n}^{t} \density(\ray(s) \, ds )\right) \density(\ray(t))\vphantom{\int_{t_n}^{t}}$\\
return $\hat{\boldsymbol C}, \hat{\boldsymbol D}$;
\end{algorithm}
\label{algo: rendering}
}
}

Algorithm 1 demonstrates our rendering procedure, starting with a ray $\mathbf r(t) = (\mathbf x_o +t\mathbf d )$ selected through spatiotemporal importance sampling. We uniformly sample points along $\mathbf r(t)$, forming a point-set $\boldsymbol{\mathcal P}u$. This set is passed to the indicator grid $\mathbf G{o}$ to obtain $\boldsymbol{\mathcal P}$, which is distributed around the target tissue. 
We initialize each point in $\boldsymbol{\mathcal P}$ with 1 (the multiplicative identity).
Subsequently, we apply bi-linear interpolation to extract features from different feature planes (\textit{e.g.}, $G_{XY}$) using corresponding dimension (\textit{e.g.}, $xy$) and update $\mathbf v$ with element-wise multiplication accordingly. These features, combined with positional encoded coordinate and time embeddings $\mathbf b(x, y, z, \tau)$, are processed by the tiny MLP $\boldsymbol\theta$ to yield color $\mathbf c_i$ and density $\mathbf \sigma_i$. Finally, volume rendering (Eq.~\ref{eq:volume_render}) is applied to calculate the color $\hat{\boldsymbol C}$ and depth $\hat{\boldsymbol D}$ of $\mathbf r(t)$ by integrating radiance and density along the ray.

\subsection{Optimization}
\label{method: Optimization}
Reconstructing surgical procedures rapidly and accurately presents significant challenges due to the inherent uncertainty and limited information. To expedite the optimization of the tiny MLP and orthogonal planes, we incorporate not only supervision from color and depth but also spatiotemporal continuity constraints. Furthermore, we devise a disentangle loss to aid in the disentanglement of static and dynamic fields. 
The following sections comprehensively demonstrate the utilization of the employed loss function in our approach.
\subsubsection{Color Loss}
We utilize captured images to optimize both the tiny MLP $\params$ and the neural plane representation $\boldsymbol{\phi} = \{\mathbf{V}_{3d}^s, \mathbf{V}_{3d}^d\}$ concurrently. We define the color loss in the following manner:
\begin{equation}
\loss{rgb}(\params, \boldsymbol{\phi}) = \sum_{i} \expect_{\ray \sim \boldsymbol\image_i}
\left[
\norm{\C(\ray) - \boldsymbol{C}^\gt_i(\ray)}_2^2
\right],
\label{eq:rgb loss}
\end{equation}
where $\boldsymbol{C}^\gt_i(\ray)$ represents the ground truth color of the ray $\ray$ passing through a pixel in image $I_{i}$.

\subsubsection{Depth Loss}
Following the methodology laid out in EndoNeRF~\cite{endonerf}, we also employ depth information generated by stereo matching to assist in the optimization of the neural scene. We define the depth loss as follows:
\begin{equation}
\loss{depth}(\params, \boldsymbol{\phi}) = \sum_{i} \expect_{\ray \sim \boldsymbol D_i}
\left[
H_{\delta}\left(\hat{\boldsymbol{D}}(\ray) - \boldsymbol{D}^\gt_i(\ray)\right)
\right],
\label{eq:gt depth loss}
\end{equation}
where the $H_{\delta}$ represent standard huber loss and $\boldsymbol{D}^\gt_i(\ray)$ represents the depth predicted by the stereo matching algorithm of the ray $\ray$.

\subsubsection{Total Variation Loss}
Inspired by \cite{fridovich2022plenoxels} and \cite{niemeyer2022regnerf}, we also leverage the total variation regularization to help optimizing our method. Specifically, the $\mathcal{L}_{TV}$ is defined as:
\begin{equation}
    \mathcal L_{TV} = \displaystyle\sum_{G \in \mathbf{V}_{3d}^{s}} \displaystyle\sum_{g \in G} \left\| \Delta^2_h (G, g) + \Delta^2_w (G, g) \right\|,
\end{equation}
where $\mathbf{V}_{3d}^{s} = \{G_{XY}, G_{YZ}, G_{XZ}\}$ denotes all space planes in Forplane. $\Delta^2_h (G, g)$ represents the squared difference between the $g$th value in plane $G:=(i, j)$ and the $g$th value in plane $G:=(i+1, j)$ normalized by the resolution, and analogously for $\Delta^2_w (G, g)$. 


\subsubsection{Time Smoothness Loss}
To robustly reconstruct deformable tissue under limited view , we further incorporate time smoothness regularization for all space-time planes. This time smoothness term, akin to total variation loss, aims to ensure similarity between adjacent frames. The formulation is as follows:
\begin{equation}
    \mathcal L_{TS} = \displaystyle\sum_{G \in \mathbf{V}_{3d}^{d}}  \displaystyle\sum_{g \in G}\left\|\Delta_t^2 (G, g)\right\|,
\end{equation}
where $\mathbf{V}_{3d}^{d} = \{G_{XT}, G_{YT}, G_{ZT}\}$ is the set of the space-time planes in Forplane. $\Delta^2_x (G, g)$ represents the difference along the time axis.


\subsubsection{Disentangle Loss}
\label{Disentangle Loss}
As discussed in Sec.~\ref{sec:introduction}, surgical scenes pose a unique challenge due to the limited viewpoints they offer. This limitation necessitates sharing information across non-sequential timesteps to enhance the reconstruction quality. Our methodology employs a static-dynamic structure to model these surgical scenes. However, this structure creates an ambiguity for the model in discerning which areas should be modeled in the static field and which ones in the dynamic field.
To tackle this issue, we commence by initializing the values in $\mathbf{V}_{3d}^{d}$ to 1, which do not influence the features in $\mathbf{V}_{3d}^{s}$ during element-wise multiplication. 
Subsequently, we optimize the model using a specialized disentangle loss function.
The $\mathcal{L}_{DE}$ is described as follows:
\begin{equation}
    \mathcal{L}_{DE} = \displaystyle\sum_{G \in \mathbf{V}_{3d}^{d}} \displaystyle\sum_{g \in G}\left\|1-g\right\|.
\label{eq: disentangle loss}
\end{equation}
With this regularization, the features of the space-time planes will tend to remain the initial value. This strategy allows us to maximize the use of the static field in modeling static scenes while also appropriately accounting for dynamic elements.


\textbf{Total Loss}: The total loss optimized during each iteration is defined as:
\begin{align}
    \mathcal{L}_{total}(\params, \mathbf{\boldsymbol{\phi}}) =& \mathcal{L}_{rgb} + \lambda_{d} \mathcal{L}_{D} + \lambda_{tv} \mathcal{L}_{TV} \notag \\  
    &+ \lambda_{ts} \mathcal{L}_{TS} + \lambda_{de} \mathcal{L}_{DE}.
\end{align}
In all experiments conducted, we set the parameters $\lambda_d = 1, \lambda_{tv} = 0.001, \lambda_{ts} = 0.05, \lambda_{de} = 0.001$, unless otherwise specified in the experimental design.

\subsection{Forplane with Monocular Inputs}
\label{method: Forplane with Monocular Inputs}
In endoscopic procedures, monocular cameras are commonly used due to their compact size, despite some laparoscopes having stereo scope cameras. 
In the context of these scopes, the reliance on stereo depth in the original method \cite{lerplane} poses limitations. To address this, our enhanced version introduces an innovative optimization method tailored for monocular scopes. This method integrates data-driven monocular geometric cues into our training pipeline more effectively. In particular,
we employ an off-the-shelf monocular depth predictor~\cite{bhat2023zoedepth} to generate a depth map $\{\bar{\boldsymbol{D}}_i\}_{i=1}^N$ for each input RGB image $\{\boldsymbol{I}_i\}_{i=1}^N$. Note that estimating absolute scale in surgical scenes is challenging; hence, depth should be treated as a relative cue. Yet, this relative depth information is beneficial even over larger distances in the image.



In light of these considerations, we introduce a monocular depth loss $\mathcal{L}_{Mono}$ to enforce consistency between the estimated depth $\hat{\boldsymbol{D}}(\ray)$ and the predicted monocular depth $\bar{\boldsymbol{D}}(\ray)$:
\begin{align}
    \mathcal L_{Mono}(\params, \boldsymbol{\phi}) &= \sum_{i} \expect_{\ray \sim \bar{\boldsymbol{D}}_i}
\left\|(\eta \hat{\boldsymbol{D}}(\mathbf{r})+\epsilon)-\bar{\boldsymbol{D}}(\mathbf{r})\right\|^{2}, \notag \\
\text{where}~\eta &= \left({\hat{\boldsymbol{D}}(\ray)}^T \hat{\boldsymbol{D}}(\ray)\right)^{-1} {\hat{\boldsymbol{D}}(\ray)}^T \bar{\boldsymbol{D}}(\ray), \notag \\
\epsilon &= \bar{\boldsymbol{D}}(\ray) - \eta \hat{\boldsymbol{D}}(\ray).
\end{align}
The $\eta$ and $\epsilon$ are the closed-form solution obtained via a least-squares criterion.
These parameters are calculated separately for each batch, as the predicted depth maps can vary in scale and shift across different batches. For monocular endoscopic sequences, we replace $\mathcal{L}_{D}$ with $\mathcal{L}_{Mono}$ during training. 
\section{experiments}
\subsection{Datasets}
\subsubsection{EndoNeRF Dataset} 
We assess the performance of Forplane using the EndoNeRF dataset~\cite{endonerf}. This dataset is a collection of typical robotic surgery videos captured via stereo cameras from a single viewpoint during in-house DaVinci robotic prostatectomy procedures. Specifically designed to capture challenging surgical scenarios characterized by non-rigid deformation and tool occlusion, the dataset serves as an ideal platform for evaluation. 
The dataset is composed of six video clips (in total 807 frames) with resolution of 512$\times$640 and lasting 4-8 seconds at a frame rate of 15 frames per second. The surgical challenges depicted in each case varies. Specifically, two cases (\textit{thin structure}, \textit{traction}) show traction on thin structures. Another two cases (\textit{pushing tissues}, \textit{pulling tissues}) depict significant manipulation of tissue through pushing or pulling. The final two cases (\textit{cutting twice}, \textit{tearing tissues}) capture the process of tissue cutting. 
These scenarios effectively illustrate the challenges inherent in managing soft tissue deformation and tool occlusion during surgery. 

\subsubsection{Hamlyn Dataset}
We evaluate our method using the public Hamlyn dataset~\cite{hamlyn1,hamlyn2}, which includes both phantom heart and in-vivo sequences captured during da Vinci surgical robot procedures. The rectified images, stereo depth, and camera calibration information are from \cite{endo-depth-and-motion}. 
To generate instrument masks, we utilize the widely-used vision foundation model, Segment Anything~\cite{kirillov2023segment}, which enables semi-automatic segmentation of surgical instruments.

The Hamlyn dataset presents a rigorous evaluation scenario as it contains sequences that depict intracorporeal scenes with various challenges, such as weak textures, deformations, reflections, surgical tool occlusion, and illumination variations. We select seven specific sequences from the Hamlyn dataset (Sequence: rectified01, rectified06, rectified08, and rectified09), each comprising 301 frames with a resolution of 480 $\times$ 640. These sequences span approximately 10 seconds and feature scenarios involving surgical tool occlusion, extensive tissue mobilization, large deformations, and even internal tissue exposure. To train and evaluate our method effectively, we divide the frames of each sequence into two sets: a training set of 151 frames and an evaluation set comprising the remaining frames. This division allows us to train our method adequately while providing a robust set for evaluation.

\begin{figure}[tbp]
  \centering
  \begin{minipage}[t]{0.5\columnwidth}
    \centering
    \includegraphics[width=0.975\textwidth]{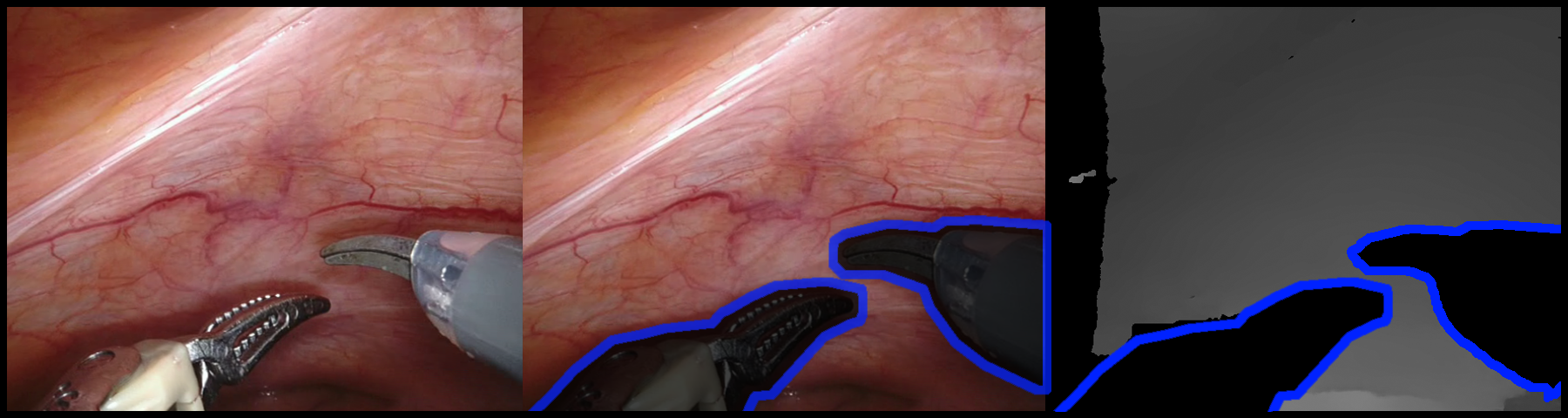}
    \label{fig:figure1}
  \end{minipage}%
  \begin{minipage}[t]{0.5\columnwidth}
    \centering
    \includegraphics[width=0.975\textwidth]{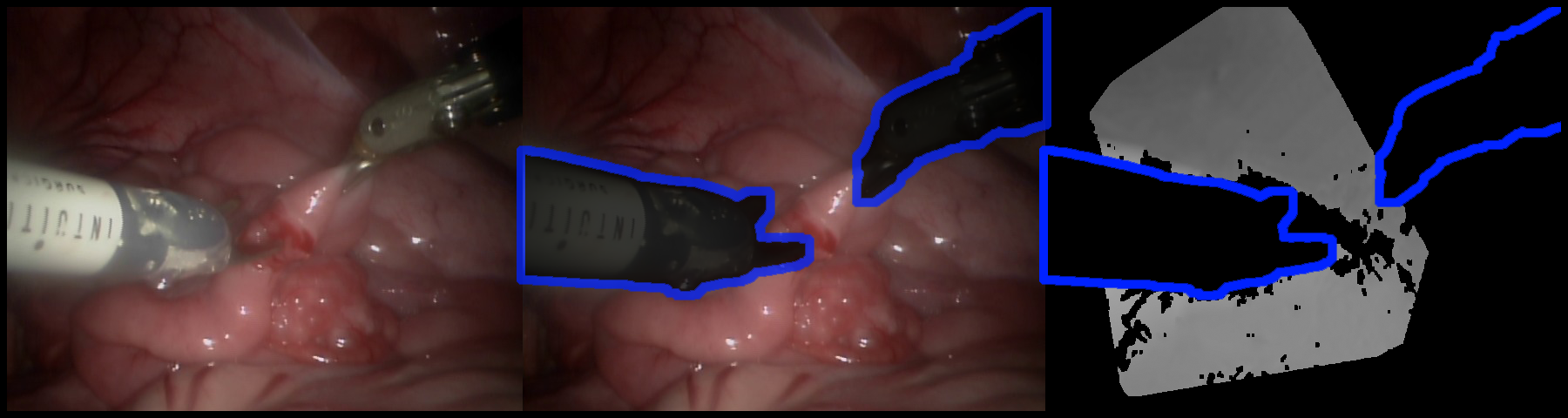}
    \label{fig:figure2}
  \end{minipage}
\caption{Visualization of datasets. The left column is from EndoNeRF dataset and the right is from Hamlyn dataset. From left to right is original image, tool mask and the binocular depth with tool mask.
}
\label{fig: dataset}
\end{figure}

\begin{figure}[tbp]
\centerline{\includegraphics[width=1.0\columnwidth]{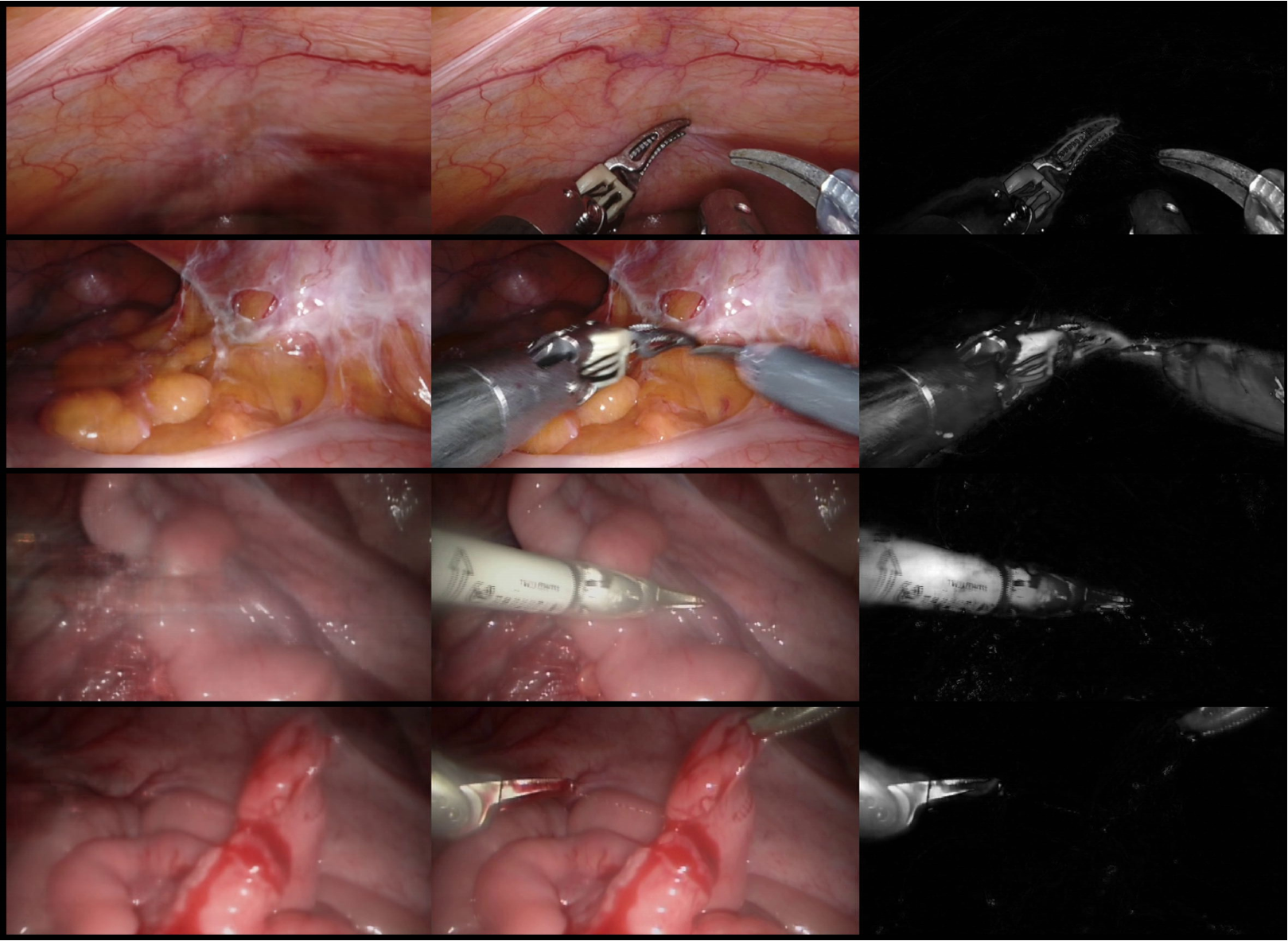}}
\vspace{-0.5em}
\caption{Reconstruction results with difference maps. From left to right, we present the image rendered by Forplane, the corresponding ground truth image, and the difference map between them. These difference maps serve as a clear visual representation of our strong ability to reconstruct deformable tissues from tool-occluded endoscopy videos.
}
\vspace{-1.5em}
\label{fig: our performance b}
\end{figure}


\begin{figure*}[htbp]
  \centering
  \begin{minipage}[b]{0.24\textwidth}
    \includegraphics[width=\linewidth]{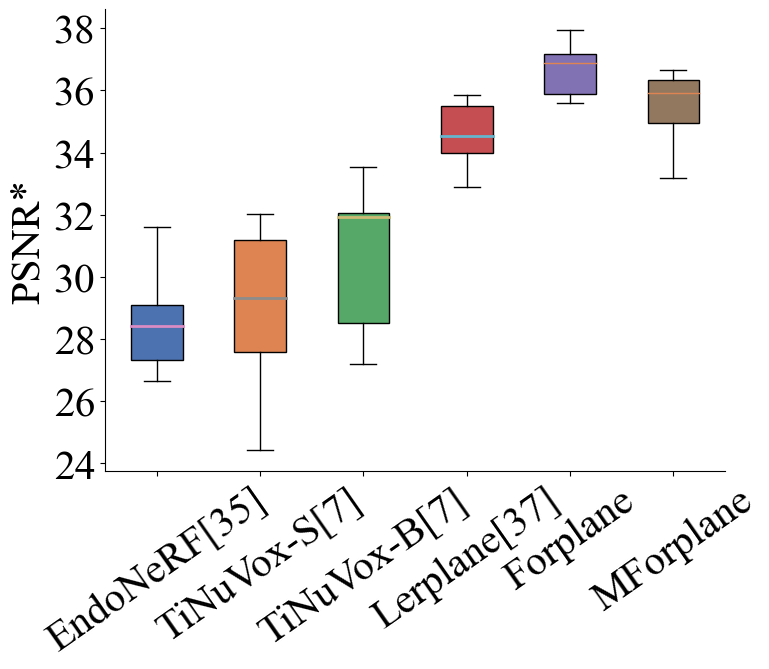}
    \label{fig:5}
  \end{minipage}%
  \hfill
  \begin{minipage}[b]{0.24\textwidth}
    \includegraphics[width=\linewidth]{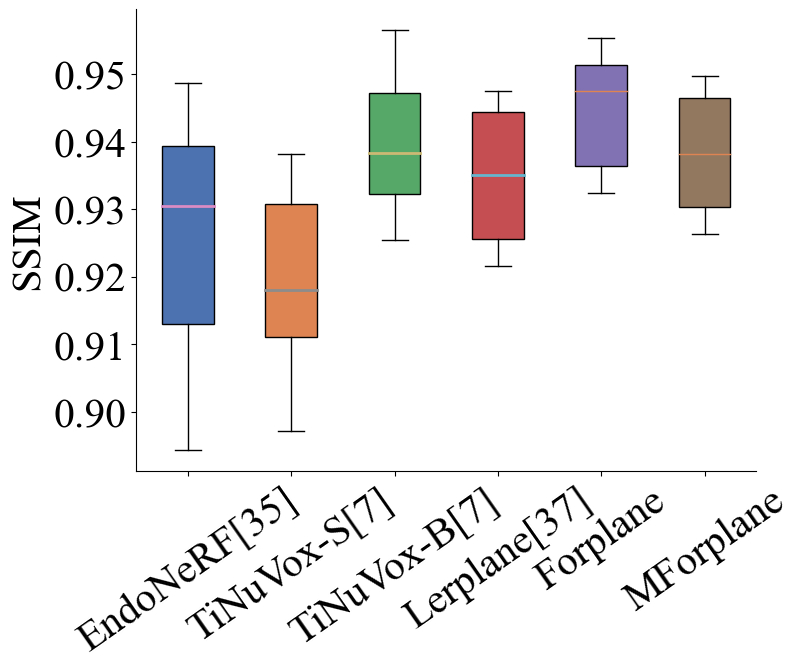}
    \label{fig:2}
  \end{minipage}%
  \hfill
  \begin{minipage}[b]{0.24\textwidth}
    \includegraphics[width=\linewidth]{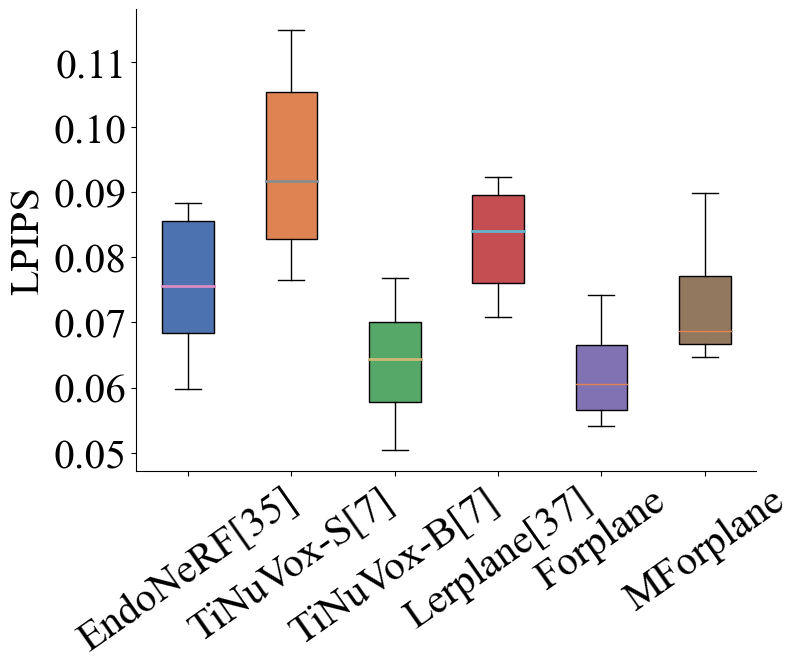}
    \label{fig:3}
  \end{minipage}%
  \hfill
  \begin{minipage}[b]{0.24\textwidth}
    \includegraphics[width=\linewidth]{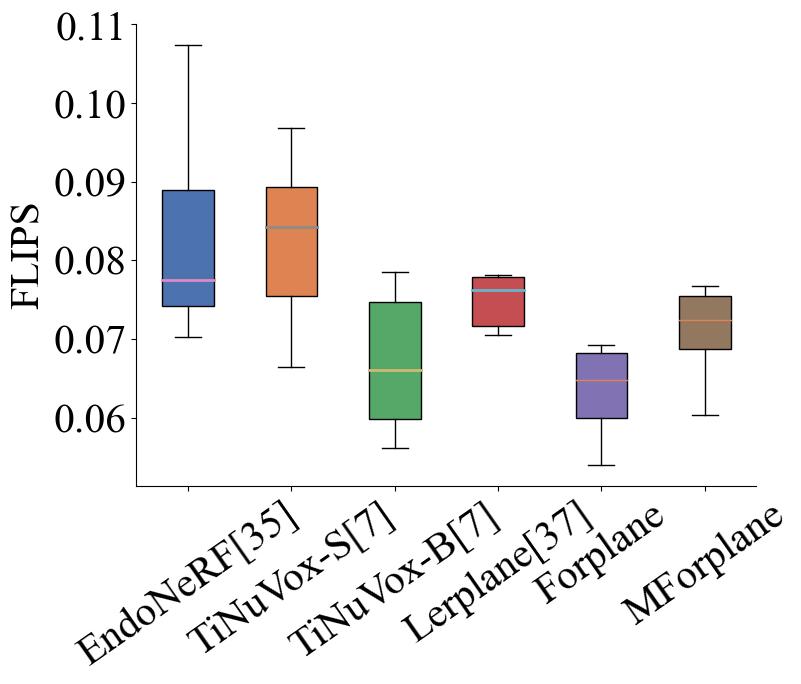}
    \label{fig:4}
  \end{minipage}%
  \caption{The Quantitative Comparison of Similarity Metrics for Reconstructed Volumes.
These figures present a quantitative comparison of similarity metrics between the input slices and the corresponding slices extracted from the reconstructed volumes. Four different similarity metrics, namely PSNR$^\star$, SSIM, LPIPS, FLIPS, are evaluated. Each figure depicts the performance of six different methods.}
  \label{fig:all}
\end{figure*}

\begin{figure*}[htp]
\centerline{\includegraphics[width=\textwidth]{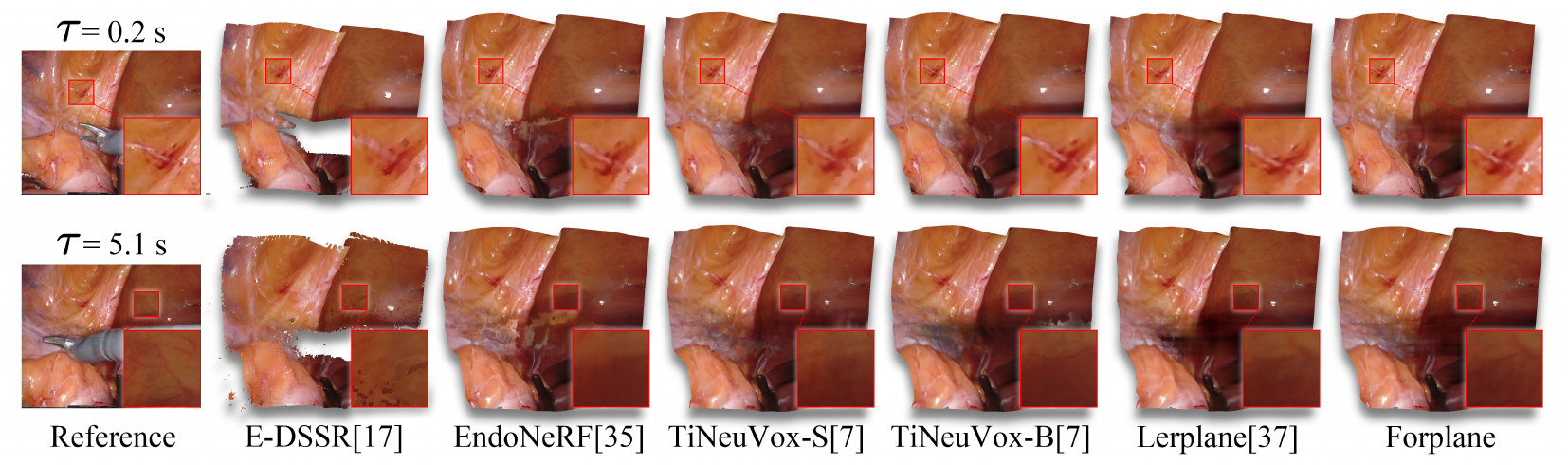}}
\caption{Results on the scene \textit{cutting twice} from EndoNeRF dataset~\cite{endonerf}. We present a comparison of tissue reconstructions from various methods. Forplane consistently shows more detailed reconstructions of the surgical procedure (\textit{e.g.} capillaries and peritoneum) compared to other methods.}
\label{fig: qualitative results}
\end{figure*}

\subsection{Baselines}
We compare our approach with three state-of-the-art methods: EDSSR \cite{edssr}, EndoNeRF \cite{endonerf} and TiNeuVox \cite{tineuvox}. 
Each of these methods represents a different type of deformable tissue reconstruction. EDSSR \cite{edssr} leverages tissue deformation fields and volumetric fusion. EndoNeRF \cite{endonerf} utilizes dynamic neural radiance fields in MLPs to represent deformable surgical scenes and optimizes shapes and deformations in a learning-based manner. We follow the standard setting in~\cite{endonerf} with 200k optimizing iterations.
TiNeuVox \cite{tineuvox} integrates optimizable time-aware voxel features for faster optmization.
Note that \cite{tineuvox} is not specified for surgical procedures, so we modify the original TiNeuVox model with tool mask-guided ray casting and stereo depth-cueing ray marching which are proposed by~\cite{endonerf}.
These methods are equipped with same hyper-parameters as~\cite{endonerf} and enables the TiNeuVox to reconstruct the dynamic tissues. We report two versions of TiNeuVox: TiNeuVox-S and TiNeuVox-B, representing the small and base versions, respectively. 
We also report Lerplane's performance~\cite{lerplane}, the previous version of Forplane. For fair comparison, we train Lerplane, TiNeuVox and Forplane with 32k iterations. 
We further report Forplane with only 9k iterations, showing significant optimization speed.


\subsection{Qualitative and Quantitative Results}
We show the quantitative results on EndoNeRF dataset in Fig.~\ref{fig:all} and all quantitative results on two datasets are summarized in Table~\ref{Tab: quantative} and Table~\ref{tab:time_speed}. All experiments were conducted on an Ubuntu 20.04 system equipped with one NVIDIA GeForce RTX 3090 GPU.
Four metrics are used for evaluation, $\textit{i.e.}$ PSNR, SSIM~\cite{ssim}, LPIPS~\cite{lpips}, FLIP~\cite{andersson2020flip}. These metrics are applied to the whole image including the masked empty areas. These areas may diminish the discernible differences between predicted and ground truth images. To address this, we introduce a modified metric called masked PSNR (PSNR*), which focuses solely on measuring the PSNR of tissue pixels, allowing for a more accurate comparison.
Apparently, Forplane outperforms other state-of-the-art methods across all evaluation metrics, demonstrating significantly superior performance. E-DSSR~\cite{edssr} achieves the fastest rendering speed due to its neural computing-light rendering procedure; however, its reconstruction quality is relatively low compared to other methods.
EndoNeRF~\cite{endonerf} shows promising reconstruction quality but requires approximately 14 hours for per-scene training. In contrast, our Forplane completes one scene learning in just 10 minutes while achieving better performance across all evaluation metrics. TiNeuVox-S~\cite{tineuvox} has a similar training time as Forplane-32k but exhibits significantly lower performance compared to our approach. TiNeuVox-B~\cite{tineuvox}, with a larger parameter quantity than TiNeuVox-S, exhibits stronger fitting ability. While better reconstruction quality is achieved on datasets with better quality (EndoNeRF dataset), the improvement diminishes on more challenging datasets (Hamlyn dataset). 
Furthermore, TiNeuVox-B accommodates a greater number of parameters, leading to increased computational time for both optimization and inference relative to TiNeuVox-S. In comparison, Forplane achieves high-quality reconstruction on both datasets with significantly faster rendering speed even with only 3 minutes optimization.
Our previous method~\cite{lerplane} exhibits performance akin to rapid optimization speeds, yet it lags in terms of rendering speed. Conversely, Forplane not only delivers superior overall quality but also boasts a rendering process that is roughly five times swifter than that of the previous version~\cite{lerplane}. This highlights the remarkable efficiency of our ray marching approach.
Furthermore, qualitative comparisons are provided in Fig.~\ref{fig: qualitative results} and the fine detailed reconstruction results are shown in Fig~\ref{fig: our performance a} and Fig.~\ref{fig: our performance b}.
The shared static field in Forplane enables better utilization of information throughout the surgical procedure, resulting in finer and more accurate details compared to other methods. Notably, Forplane achieves these improvements with less training time and faster rendering speed.

\begin{table*}[htp]
\centering
\caption{Mean values of quantitative metrics for different models on the two datasets (standard deviation in parentheses). \\$\uparrow$ indicates that higher values indicate higher accuracy, and vice versa.}
\label{Tab: quantative}
\begin{tabular}{l|ccccc}
\hline
Methods & PSNR$\uparrow$ & $\text{PSNR}^\star\uparrow$ & SSIM$\uparrow$ & LPIPS$\downarrow$ & FLIP$\downarrow$ \\
\hline
\multicolumn{6}{c}{EndoNeRF Dataset~\cite{endonerf}} \\
\hline
E-DSSR\cite{edssr} & 13.398 (1.270) & 12.997 (1.232) & 0.630 (0.057) & 0.423 (0.047) & 0.426 (0.056) \\
EndoNeRF\cite{endonerf} & 29.477 (1.886) & 28.560 (1.782) & 0.926 (0.021) & 0.080 (0.019) & 0.083 (0.014) \\
TiNeuVox-S\cite{tineuvox} & 29.913 (3.069) & 28.980 (2.862) & 0.919 (0.015) & 0.094 (0.015) & 0.083 (0.011) \\
TiNeuVox-B\cite{tineuvox} & 31.601 (2.855) & 30.668 (2.697) & \underline{0.940} (0.012) & \underline{0.064} (0.010) & \underline{0.067} (0.009) \\
Lerplane\cite{lerplane} & \underline{35.504} (1.161) & \underline{34.575} (1.129) & 0.935 (0.011) & 0.083 (0.009) & 0.075 (0.010) \\
Forplane-9k & 33.374 (1.074) & 32.435 (1.165) & 0.907 (0.014) & 0.127 (0.019) & 0.093 (0.007) \\
Forplane-32k & \textbf{37.306} (1.406) & \textbf{36.367} (1.521) & \textbf{0.945} (0.010) & \textbf{0.062} (0.008) & \textbf{0.063} (0.006) \\
\hline
\multicolumn{6}{c}{Hamlyn Dataset~\cite{hamlyn1, hamlyn2}} \\
\hline 
E-DSSR\cite{edssr} & 18.150 (2.571) & 17.337 (2.402) & 0.640 (0.060) & 0.393 (0.066) & 0.259 (0.065) \\
EndoNeRF\cite{endonerf} & 34.879 (1.784) & 34.066 (1.717) & 0.951 (0.011) & \underline{0.071} (0.017) & 0.070 (0.012) \\
TiNeuVox-S\cite{tineuvox} & 35.277 (1.682) & 34.464 (1.501) & \underline{0.953} (0.014) & 0.085 (0.029) & \underline{0.067} (0.016) \\
TiNeuVox-B\cite{tineuvox} & 33.764 (2.047) & 32.951 (1.974) & 0.942 (0.020) & 0.146 (0.061) & 0.078 (0.022) \\
Lerplane\cite{lerplane} & 32.455 (2.247) & 31.629 (2.064) & 0.935 (0.021) & 0.124 (0.041) & 0.098 (0.028) \\
Forplane-9k & \underline{35.301} (2.241) & \underline{34.475} (2.076) & 0.945 (0.018) & 0.093 (0.035) & 0.074 (0.020) \\
Forplane-32k & \textbf{37.474} (2.401) & \textbf{36.647} (2.232) & \textbf{0.960} (0.014) & \textbf{0.058} (0.025) & \textbf{0.059} (0.017) \\
\hline
\end{tabular}
\end{table*}

\begin{table}[tbp]
\centering
\caption{Training Time and Test Speed for Different Methods}
\label{tab:time_speed}
\begin{tabular}{l|cc}
\hline
Methods & Train Time & Test Speed \\
\hline
E-DSSR\cite{edssr} & 13 mins & \textbf{28.0} fps \\
EndoNeRF\cite{endonerf} & $>$10 hours & 0.11 fps \\
TiNeuVox-S\cite{tineuvox} & 12 mins & 0.56 fps \\
TiNeuVox-B\cite{tineuvox} & 90 mins & 0.18 fps \\
Lerplane\cite{lerplane} & 10 mins & 0.38 fps \\
Forplane-9k & ~3 mins & \underline{1.73} fps \\
Forplane-32k & 10 mins & \underline{1.73} fps \\
\hline
\end{tabular}
\vspace{-1.5em}
\end{table}

Experimental results indicate that Forplane excels in computational speed and reconstruction quality, positioning it as a promising tool for future clinical and intraoperative applications in surgery. 
Through its pioneering method that bolsters optimization without sacrificing the quality of reconstruction, Forplane stands poised to transform the landscape of surgical procedures.

\subsection{Ablation Study}
In this section, we perform a series of experiments to validate the effectiveness of our proposed methods. The experiments are conducted on the EndoNeRF dataset~\cite{endonerf} with 9k iterations, and the reported metric values are averaged.

\noindent \textbf{Sampling Strategy}
We compare our proposed spatiotemporal importance sampling strategy with two other sampling strategies: Naive Sampling, which avoids tool masks and assigns equal weights to all pixels, and EndoNeRF Sampling, which assigns higher probabilities to highly occluded areas as described in \cite{endonerf}.

\noindent \textbf{Encoding Method}
We compare our coordinate-time encoding method with two methods: Dummy Encoding, which replaces all the coordinate-time encoded parameters with a constant value to preserve the parameter quantity of the tiny MLP, and Direction Encoding, where one-blob positional encoding is applied to the view direction and the encoded parameters are passed to the tiny MLP, as described in \cite{nerf}.

\noindent \textbf{Space-time Disentangle}
As detailed in Sec.~\ref{Disentangle Loss}, we use a specialized initialization for $\mathbf{V}_{3d}^d$ and pair it with $\mathcal{L}_{DE}$ to enhance the decomposition between $\mathbf{V}_{3d}^d$ and $\mathbf{V}_{3d}^s$. For comparison, we also trained a version of Forplane without these adjustments, termed w/o Disentangle.

\noindent \textbf{Ray Marching}
In this improved version, we introduce an efficient ray marching algorithm designed to accurately generate spatial-temporal points. To evaluate the effectiveness of this algorithm, we conducted a comparative analysis against two existing methods: the sample-net method (termed Sample-Net) and the stereo depth-cueing ray marching method (termed Depth-Cueing). Sample-Net, as proposed in the original version of our work~\cite{lerplane}, utilizes a simplified representation of the 4D scene, aiming to enhance the accuracy of point sampling. In contrast, the Depth-Cueing method, introduced in EndoNeRF~\cite{endonerf}, employs Gaussian transfer functions informed by stereo depth data to guide the sampling of points near tissue surfaces, aiming to optimize the rendering of fine details.

\begin{table}[tbp]
\centering
\caption{Mean values of evaluation metrics for different ablation models on the EndoNeRF~\cite{endonerf} dataset.}
\label{tab: ablation}
\begin{tabular}{l|ccccc}
\hline
Methods & PSNR$\uparrow$ & SSIM$\uparrow$ & LPIPS$\downarrow$ & FLIP$\downarrow$ \\
\hline
Naive Sampling  & 33.146 & 0.906 & 0.135 & 0.094 \\
EndoNeRF Sampling  & 33.019 & 0.906 & 0.135 & 0.093 \\
Dummy Encoding  & 33.055 & 0.905 & 0.129 & 0.094 \\
Direction Encoding  & 33.062 & 0.904 & 0.133 & 0.095 \\
W/o Disentangle & 18.015 & 0.869 & 0.302 & 0.349 \\
\hline
Full Model  & 33.374 & 0.907 & 0.127 & 0.093 \\
\hline
\end{tabular}
\end{table}

\begin{figure*}[tbp]
\centerline{\includegraphics[width=\textwidth]{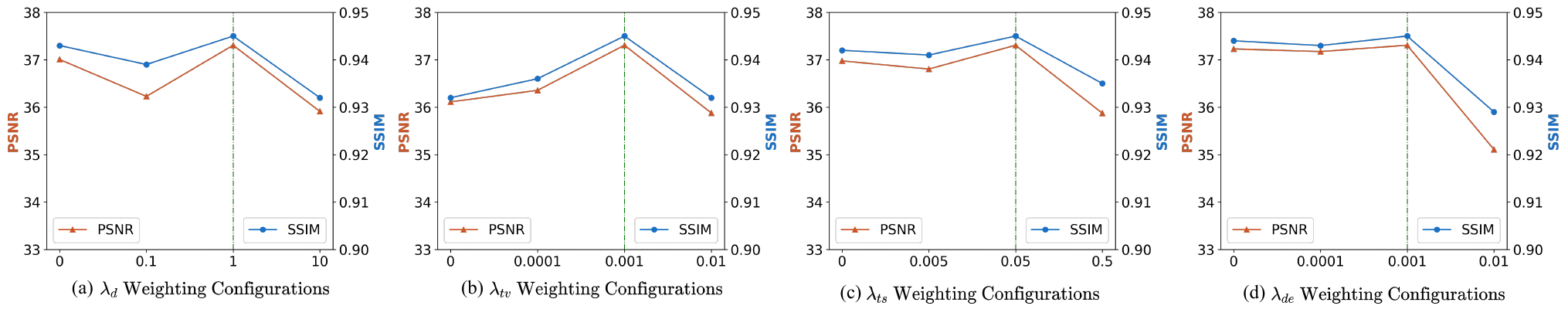}}
\caption{Comparative reconstruction results with varied weighting parameters. This figure illustrates the performance impact of different weighting configurations on the reconstruction process, with the horizontal axis represented in a logarithmic scale. Fig.~9 (a) shows the performance of Forplane with different $\lambda_d$. Fig.~9 (b) shows $\lambda_{tv}$, Figure (c) shows $\lambda_{ts}$ and Figure (d) shows $\lambda_{de}$.
}
\label{fig: lambda}
\end{figure*}

\begin{table}[tbp]
\centering
\caption{Evaluation metrics for different ray marching strategy on the EndoNeRF~\cite{endonerf} dataset.}
\label{tab: ablation ray marching}
\begin{tabular}{l|ccccc}
\hline
Methods & PSNR$\uparrow$ & SSIM$\uparrow$ & LPIPS$\downarrow$ & Test Speed$\uparrow$ \\
\hline
Depth-Cueing &	31.325 &	0.897 &	0.132 & 1.95 fps \\
Sample-Net	&32.889	&0.901	&0.126	&0.38 fps \\
Ours	&33.374&	0.907&	0.127&	1.73 fps \\
\hline
\end{tabular}
\end{table}

The results of the ablation study are summarized in Table~\ref{tab: ablation} and Table~\ref{tab: ablation ray marching}. Our spatiotemporal importance sampling achieves superior quality compared to the other sampling strategies, confirming its effectiveness. Regarding encoding methods, our coordinate-time encoding outperforms the others, while the performance of Dummy and Direction Encoding is similar, validating of our hypothesis in Section \ref{method: Coordinate-Time Encoding}.
Forplane w/o Disentangle shows notable degradation, primarily from optimization ambiguities. Lacking clear guidance on decomposition, the network struggles to distinguish between static and dynamic elements given the restricted viewpoints.
The results presented in Table~\ref{tab: ablation ray marching} indicate that the depth-cueing method yields suboptimal performance, primarily due to the reliance on noisy and incomplete stereo depth data, as depicted in Fig.~\ref{fig: dataset}. Additionally, its inapplicability for temporal interpolation further limits its utility due to the absence of stereo depth in certain contexts. Conversely, while the sample-net method is versatile, it leads to increased computational demands during inference, negatively impacting test speed.
Our efficient ray marching method effectively mitigates these issues, striking a balance between rendering speed and reconstruction quality. It outperforms the depth-cueing method in terms of PSNR and SSIM, and also surpasses the sample-net method in inference speed, thus demonstrating its potential for real-time application scenarios.

\begin{table}[tbp]
\centering
\caption{The performance between Forplane and MForplane. The time row represents for optimization time.}
\label{tab: monocular}
\begin{tabular}{l|ccccc}
\hline
Methods & PSNR$\uparrow$ & $\text{PSNR}^\star\uparrow$ & SSIM$\uparrow$ & LPIPS$\downarrow$ & FLIP$\downarrow$ \\
\hline
\multicolumn{6}{c}{EndoNeRF Dataset~\cite{endonerf}} \\
\hline
Forplane & 37.306 & 36.367 & 0.945 & 0.062 & 0.063 \\
MForplane  & 36.403 & 35.464 & 0.938 & 0.073 & 0.071 \\
\hline
\multicolumn{6}{c}{Hamlyn Dataset~\cite{hamlyn1, hamlyn2}} \\
\hline
Forplane & 37.474 & 36.647  & 0.960  & 0.058  & 0.059  \\
MForplane  & 37.034 & 36.208 & 0.959 & 0.060 & 0.063 \\
\hline
\end{tabular}
\end{table}

\subsection{Hyperparameters}
As delineated in Section~\ref{method: Optimization}, we set the $\lambda$ parameters to a specific weighting configuration across all experiments. This subsection is dedicated to conducting a comprehensive analysis to discern the effects of various hyperparameters on Forplane's performance. To this end, we train Forplane using an array of $\lambda$ weightings on the EndoNeRF dataset~\cite{endonerf}, extending across 32k iterations. The results, depicted in Fig.~\ref{fig: lambda}, suggest that Forplane exhibits considerable robustness to variations in loss weighting configurations, only excessively large weightings could potentially impair performance.

\subsection{Monocular Reconstruction}
We show the performance of MForplane and Forplane on two datasets in Table.~\ref{tab: monocular}. By incorporating depth priors from monocular depth estimation networks, MForplane achieves fast reconstruction of dynamic tissues in monocular endoscopic videos. However, when compared to Forplane, we observe a slight decrease in the reconstruction quality of MForplane (\textit{e.g.} 1.81\% in PSNR, 0.4\% in SSIM).
Furthermore, there is a slight increase in the training time (from 10 minutes to 12 minutes), primarily attributed to the additional computational workload involved in calculating the monocular depth loss. Nevertheless, it is important to note that the ability to achieve fast reconstruction of dynamic tissues using one monocular endoscopic video holds significant promise for various applications in medical imaging. Therefore, the minor reduction in quality is deemed acceptable given the enormous potential and practicality of MForplane.

\subsection{The Choice of MLP Structure}
As mentioned in Section \ref{method: Neural Forplane Representation for Deformable Tissues}, our fast orthogonal plane representation reduces the computational load on the MLP, allowing for a more efficient MLP architecture in complex surgical reconstruction tasks. To validate this approach, we conducted an experimental comparison where our tiny MLP (2 fully-connected ReLU layers with 64 channels) was replaced with the more elaborate MLP structure (8 fully-connected ReLU layers with 256 channels) utilized by EndoNeRF~\cite{endonerf}. The experiments are conducted on the EndoNeRF dataset with 9k iterations, and the reported metric values are averaged. The experimental results, detailed in Table~\ref{tab: ablation MLP}, indicate that the use of a larger MLP configuration does not significantly enhance reconstruction quality but results in a substantial increase in training duration for the same iteration count. These outcomes strongly corroborate our proposition that a streamlined MLP architecture contributes to enhanced optimization and inference efficiency, underscoring the effectiveness of our proposed method.

\begin{table}[tbp]
\centering
\caption{Mean values of evaluation metrics for different MLP structure on the EndoNeRF~\cite{endonerf} dataset.}
\label{tab: ablation MLP}
\begin{tabular}{l|cccc}
\hline
Methods & PSNR$\uparrow$ & SSIM$\uparrow$ & Train Time$\downarrow$ & Test Speed$\uparrow$ \\
\hline
Large MLP  & 34.285 & 0.913 & 23 mins & 0.75 fps \\
Tiny MLP  & 33.374 & 0.907 & ~3 mins & 1.73 fps \\
\hline
\end{tabular}
\end{table}

\begin{figure}[htbp]
\centerline{\includegraphics[width=0.975\columnwidth]{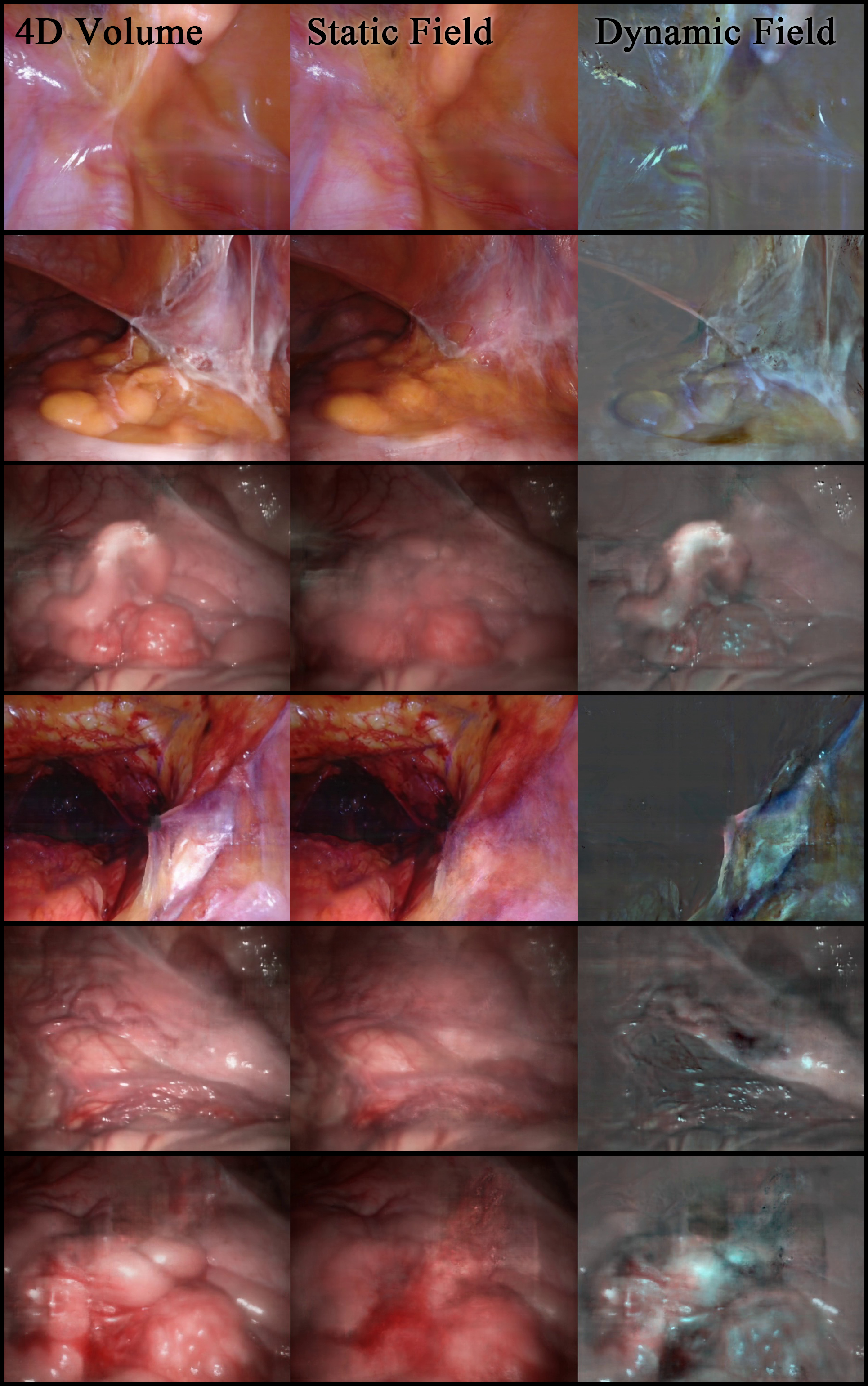}}
\caption{Visualization of static and dynamic fields. 
The visualizations provided demonstrate the decomposition of static and dynamic fields in surgical procedures, showcasing the effectiveness of utilizing scene context within Forplane. The first image of each row shows the result of full 4D volume. The second image corresponds to the static field, while the third image portrays the dynamic field, which is re-normalized to [0, 1] for optimal visualization.}
\label{fig: decomposition}
\end{figure}

\subsection{Disentangle Static and Dynamic Field}
We propose an effective method for visualizing the static and dynamic fields in surgical scenes, showcasing the efficacy of our factorization approach.
Our approach utilizes element-wise multiplication for feature aggregation across different feature planes, as outlined in Eq.~\ref{eq:interpolate}. 
By setting all values in the dynamic field to 1 and performing inference, we obtain the network's output as the rendered result of the static field. Similarly, setting all values in the static field to 1 provides the rendered result of the dynamic field. 
The time-space decomposition is clearly illustrated in Fig.~\ref{fig: decomposition}, offering a compelling demonstration of the effectiveness of our proposed structure. 
This property offers valuable insights into tissue examination and procedural understanding during surgery, with potential applications in virtual surgery and intraoperative utilization.

\section{Discussion}
Forplane exhibits SOTA performance in rapidly reconstructing high-quality deformable tissues within both monocular and binocular endoscopic videos. This significant advancement is attributed to a series of critical observations and methodological innovations:
1) Combination of Implicit and Explicit Representations: While implicit representations excel in detailed scene restoration, the computational demands are substantial. Conversely, explicit representations, though computationally efficient, often lack detail. Forplane integrates the strengths of both implicit and explicit representations, thereby achieving efficient training and inference, alongside high-quality reconstruction.
2) Selective Focus on Significant Regions: Acknowledging the varying significance of different tissues, Forplane strategically focuses on critical regions. This targeted approach accelerates the optimization process by allocating computational resources to areas of the scene with the most impact on overall reconstruction quality.
3) Utilization of Temporal Information: Given the limited viewpoint in endoscopic environments, Forplane effectively leverages temporal data. This utilization of temporal information is key in enhancing the accuracy of the reconstruction, particularly in dynamically changing surgical scenes.

Despite its notable achievements, Forplane has areas that warrant further development. First, the MForplane variant, designed to operate without stereo depth, still depends on manually generated masks for surgical instrument identification. This process is both challenging and labor-intensive, highlighting the need for advancements in automated instrument labeling and tracking.
Secondly, the rendering speed of Forplane, although significantly faster than its predecessor (approximately 5 times quicker at 1.73 fps on a single RTX3090 GPU), does not yet meet the demands for real-time intraoperative assistance. This limitation underscores the necessity for further improvements in rendering speed.
Lastly, the functionality of Forplane is presently focused on the reconstruction of deformable tissues. To fully support comprehensive surgical training for robotic systems, a more detailed analysis of surgical procedures is required. This includes not only deformable tissue reconstruction but also the intricate dynamics of surgical instruments, environmental lighting conditions, and camera motion.

\section{Conclusion}
In this paper, we propose a fast dynamic reconstruction framework, Forplane, targeted on deformable tissues during surgery. 
Forplane represents surgical procedures with orthogonal neural planes and incorporates advancements in pixel sampling, spatial sampling, and information enhancement to improve rendering quality, accelerate optimization and inference. 
Evaluations on two in vivo datasets demonstrate the superior performance of Forplane, requiring less training time and offering faster inference speed compared to other methods.
In addition, Forplane effectively handles both monocular and binocular endoscopy videos, maintaining nearly identical high-quality reconstruction performance.
We firmly believe that Forplane holds substantial promise for intraoperative applications among surgery.

{\small
\bibliographystyle{ieee_fullname}
\bibliography{egbib}
}

\end{document}